\documentclass[times,twocolumn,final,authoryear]{elsarticle}
\usepackage{hyperref}
\hypersetup{
    colorlinks=true, 
    linkcolor=blue, 
    filecolor=magenta,      
    urlcolor=cyan, 
    citecolor= blue 
    pdftitle={Overleaf Example},
    pdfpagemode=FullScreen,
    }

\usepackage{ycviu}
\usepackage{framed,multirow}

\usepackage{amssymb}
\usepackage{latexsym}

\usepackage{url}
\usepackage{xcolor}
\definecolor{newcolor}{rgb}{.8,.349,.1}

\usepackage[edges]{forest}
\forestset{
  qtree/.style={
    baseline,
    for tree={
      parent anchor=south,
      child anchor=north,
      align=center,
      inner sep=1pt,
    }}}

\begin{document}
\begin{frontmatter}
\title{Image Amodal Completion: A Survey}

\author[1]{Jiayang Ao}
\author[2]{Qiuhong Ke}
\author[1]{Krista A. Ehinger}
\address[1]{The University of Melbourne, Parkville, 3010, Australia}
\address[2]{Monash University, Clayton, 3800, Australia}

\begin{abstract}
Existing computer vision systems can compete with humans in understanding the visible parts of objects, but still fall far short of humans when it comes to depicting the invisible parts of partially occluded objects. Image amodal completion aims to equip computers with human-like amodal completion functions to understand an intact object despite it being partially occluded. The main purpose of this survey is to provide an intuitive understanding of the research hotspots, key technologies and future trends in the field of image amodal completion. Firstly, we present a comprehensive review of the latest literature in this emerging field, exploring three key tasks in image amodal completion, including amodal shape completion, amodal appearance completion, and order perception. Then we examine popular datasets related to image amodal completion along with their common data collection methods and evaluation metrics. Finally, we discuss real-world applications and future research directions for image amodal completion, facilitating the reader's understanding of the challenges of existing technologies and upcoming research trends.
\end{abstract}

\end{frontmatter}

\section{INTRODUCTION}

Amodal completion is the ability to perceive the whole object even though it is partially occluded \citep{kanizsa1979organization}. For example, when a dog's body is hidden in the grass, humans still see the dog as a complete animal and can imagine the appearance of its occluded parts. This natural ability of the human visual system allows humans to easily fill in the occluded appearance of invisible objects, distinguish the boundaries between different objects and determine the layering order of occluders and occludees. This ability to perceptually "fill in" the invisible parts of the scene is important, because it helps humans to perceive visual scene input consisting of fragmented, incomplete and disordered objects into a scene consisting of coherent, complete and continuous objects in everyday scenes \citep{chen2016amodal}.

While existing computational models can compete with humans for object recognition when objects are occlusion-free, they are much less robust than humans when dealing with partially occluded objects \citep{kortylewski2020compositional,DBLP:conf/cogsci/ZhuTYPP19}. Humans are able to perceive an entire target object through only part of its area, and this rapid recognition and spatial range awareness of an occluded object help to  reliably predict its next move \citep{ling2020variational}. When humans see half a wheel pop up around a corner, people can tell that a vehicle is about to come and therefore choose to stop and wait for the vehicle to pass first. Such everyday scenes are effortless for humans to discern, but they are a huge challenge for computer vision systems. The ability of amodal completion thus enables computers to be more human-like when processing occlusion.

Amodal completion is also of great benefit for many computer vision tasks. In particular, autonomous vision systems applied in reality must perform similar reasoning for occluded objects to guarantee operational safety and reliability. For robotic grasping systems, the ability to infer the entire structure of an occluded object from its visible parts allows the robot to directly grasp and manipulate unseen objects in cluttered scenes. For self-driving, the rapid identification of an object and its complete spatial extent from local areas of the object in a complex scene helps predict more accurately what is likely to happen in the short future and thus plan accordingly.

\begin{figure*}[h] 
\centering 
\includegraphics[width=0.9\textwidth]{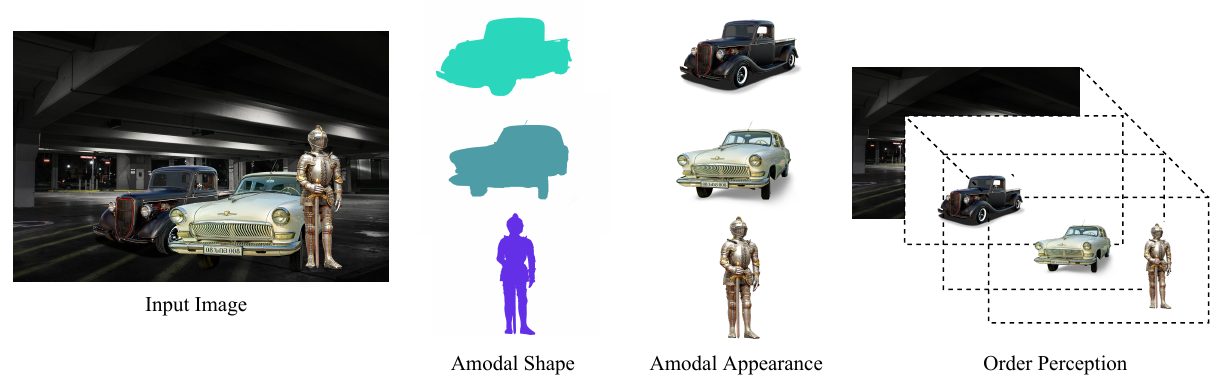} 
\caption{For an input image with multiple occlusions, the amodal completion ability allows to complete the shape and fill in the appearance of the hidden parts of objects, as well as to perceive the order between overlapping objects.}
\label{Fig.main1} 
\end{figure*}

Despite the importance of amodal completion for real-world applications, most traditional research in image understanding has ignored invisible regions and focused only on visible regions. For example, object tracking methods may down weight occluded parts because occlusions are hypothesised to be uncommon \citep{milan2016mot16}, while object recognition methods may simply exclude highly occluded objects from the evaluation \citep{russakovsky2015imagenet}. These methods of ignoring occlusions lead to failures for downstream tasks such as 3D reconstruction due to their poor performance in cases of severe occlusion \citep{Reddy_2022_CVPR}. In fact, in real-life scenes, we see more partially occluded objects than fully visible ones. The infinite possibilities of occlusion patterns between objects mean that the shape and appearance of the occluded parts are completely unconstrained, even if a priori knowledge of the object class is available in a given environment (e.g., pedestrians on a busy street). The numerous possible occlusion patterns in complex and cluttered natural environments pose a great challenge to the recognition, tracking and measurement capabilities of computer vision systems. This is why it is essential for computer vision systems to mimic the amodal perception ability of humans.

Amodal completion has recently attracted research attention with the increasing demands on the perception ability of computer vision systems, and as a result, there has been a growing body of work in this area. However, in contrast to the extensive and mature work on the visible part of the image, research on the amodal completion of images by computer vision systems is still in its infancy. A survey is therefore urgently needed to help researchers gain an effective understanding of the current state of development in this growing field.

Nevertheless, there is a lack of review studies focusing on image amodal completion. A survey of image segmentation using deep learning is provided in \citep{minaee2021image}, but it does not have a specific discussion of segmentation algorithms for invisible parts of objects. Other reviews on reasoning about the invisible parts of objects focus on specific domains, such as the automotive environment \citep{gilroy2019overcoming} or human faces \citep{zhang2018facial}. A recent survey discusses occlusion handling in generic object detection from still images \citep{saleh2021occlusion}, but the occlusion handling problem is only part of the amodal completion task. This is because amodal completion is not only able to determine the occlusion relationships between objects, but also to infer the shape and appearance of the occluded parts. An earlier article discusses amodal volume completion for 3D visual tasks \citep{breckon2005amodal}, while we attempt to comprehensively investigate amodal completion from single 2D images.

The main contributions of this paper are summarized as follows:
\begin{itemize}
\item We systematically review amodal completion based on a single image. We present a novel taxonomy that organizes various systems in terms of the input-output of tasks and technical perspectives to provide a comprehensive insight into image amodal completion.
\item We survey the latest research and public datasets on image amodal completion to 2022 and identify real-world applications that will enable future researchers to apply image amodal completion techniques.
\item We discuss the challenges of image amodal completion, highlight open issues that need more attention, and suggest possible avenues for future research.
\end{itemize}

\section{TAXONOMY of PROBLEMS}

The amodal completion process of the visual system generally addresses three key problems, as illustrated in Fig.\ref{Fig.main1}. The first is to complete the full shape of each partially occluded object, usually represented by a binary segmentation mask for the entire instance, including its invisible area. The second is to infer the plausible appearance of the invisible parts of an object from its visible parts, i.e., generating the RGB values of the hidden pixels of the object. Another key issue arising from the first two problems is identifying the correct order between overlapping objects. Object perception, i.e., inferring the order between objects, is therefore employed in some algorithms to assist in solving the previous two problems. These three problems have been studied individually and jointly in recent work. A list of the main tasks we will discuss regarding image amodal completion is summarised in Table \ref{tab:main_works}. In this section, we will discuss representative works for each task and analyse the strengths and challenges of these works.

\begin{table*}[!h]
\renewcommand\arraystretch{1.2}
\centering
\footnotesize 
\caption{Summary of the main works we discuss on image amodal completion. Tasks: A: amodal appearance. O: order perception. S: amodal shape. Input: IMG: image. MBB: modal (or visible) bounding box. MM: modal segmentation mask. OB: occlusion boundary. Output: ABB: amodal bounding box. AM: amodal segmentation mask. ASL: amodal scene layout. CC: completion curve. LO: layer order. OO: occlusion order. RGB: intact RGB object or/and background. Sup.: Supervision. N/A: not applicable. RL: reinforcement learning. SL: supervised learning. SSL: self-supervised learning. UL: unsupervised learning. WL: weakly supervised learning. APP.: the application scenarios of these methods. Objects: common objects in a natural context. }

\begin{tabular}{|l|l|l|l|l|l|l|}
\hline
Tasks & Paper  & Methods  & Input  & Output & Sup. & APP.  
\\ \hline
S  & \cite{kimia2003euler} & Geometry-based & IMG   & CC   & N/A &Simple shapes   \\ \hline
S  & \cite{walton2008improved}  & Geometry-based & IMG  & CC & N/A & Simple shapes \\ \hline
S  & \cite{silberman2014contour} & Geometry-based & IMG  & CC  & N/A & Indoor scenes \\ \hline
S  & \cite{kar2015amodal} & CNN-based & IMG, MBB & ABB & SL & Objects   \\ \hline
S  & \cite{li2016amodal}  & CNN-based (IBBE)
& IMG, MBB & ABB, AM & SL & Objects    \\ \hline
S  & \cite{lin2016computational}   & Geometry-based & IMG  & CC  & N/A & Simple shapes \\ \hline
S  & \cite{kihara2016shadows} & Shape Boltzmann Machine  & IMG, MM & AM & SL & Objects \\ \hline
O, S  &  \cite{zhu2017semantic}  & CNN-based (AmodalMask) & IMG         & AM, OO      & SL & Objects         \\ \hline
A, O, S &  \cite{ehsani2018segan}  & \begin{tabular}[c]{@{}l@{}}CNN + cGAN (SeGAN)\end{tabular}   & IMG, MM     & AM, LO, RGB & WL & Indoor scenes                \\ \hline
S  & \cite{qi2019amodal}     & CNN-based (ASN)   & IMG         & AM          & SL & Vehicles   \\ \hline
O, S  &  \cite{zhang2019learning} & CNN-based &IMG & AM, LO & SL & Objects \\ \hline
S  &  \cite{follmann2019learning}         & \begin{tabular}[c]{@{}l@{}}CNN-based (ORCNN)\end{tabular}               & IMG         & AM          & SL & Objects \\ \hline
O, S  &  \cite{purkait2019seeing}             & CNN-based & IMG         & AM, LO      & SL & Indoor scenes \\ \hline
S  & \cite{hu2019sail} & CNN-based & IMG         & AM          & SL & Objects   \\ \hline
A  &  \cite{burgess2019monet}              & \begin{tabular}[c]{@{}l@{}}VAE + Attention (MONet)\end{tabular}          & IMG         & MM, RGB     & UL & Toy dataset \\ \hline
A  & \cite{greff2019multi}                  & \begin{tabular}[c]{@{}l@{}}VAE-based (IODINE)\end{tabular}               & IMG         & MM, RGB     & UL & Toy dataset \\ \hline
A, S & \cite{yan2019visualizing}   & GAN-based  &  IMG    & AM, RGB     & WL & Vehicles  \\ \hline
A, O, S  &\cite{Papadopoulos_2019_CVPR} & \begin{tabular}[c]{@{}l@{}}GAN-based (PizzaGAN)\end{tabular} &  IMG  & AM, LO, RGB & WL & Pizza  \\ \hline
A, O, S & \cite{dhamo2019object}  & CNN-based   & IMG  & AM, LO, RGB & WL & Indoor scenes  \\ \hline
A, O  & \cite{dhamo2019peeking} & CNN + GAN &  IMG         & LO, MM,  RGB & SL & Indoor scenes \\ \hline
S  & \cite{wang2020robust} & CompositionalNet (Context-aware) & IMG & ABB &  SL & Vehicles  \\ \hline
A, O, S  & \cite{zhan2020self} & PCNet &  IMG, MM     & AM, OO, RGB & SSL & \begin{tabular}[c]{@{}l@{}}Objects, Vehicles\end{tabular}  \\ \hline
A, S  &  \cite{ling2020variational} & \begin{tabular}[c]{@{}l@{}}VAE-based (Amodal-VAE)\end{tabular}           &IMG, MM     & AM, RGB     & WL & Vehicles \\ \hline
A  & \cite{engelcke2020genesis}         & \begin{tabular}[c]{@{}l@{}} Probabilistic Generative Model\\(GENESIS)\end{tabular} &  IMG & MM, RGB & RL & Toy dataset  \\ \hline
A  & \cite{locatello2020object}         & \begin{tabular}[c]{@{}l@{}}CNN + Attention \\(Slot Attention)\end{tabular} &  IMG         & MM, RGB    & UL & Toy dataset  \\ \hline
S  &\cite{mani2020monolayout} & Adversarial learning (MonoLayout) &  IMG    & ASL  & SSL & Road scenes \\ \hline
S  & \cite{narasimhan2020seeing}   & Learning-based   &  IMG & ASL  & RL & Indoor scenes  \\ \hline
O, S  &\cite{ke2021deep} & \begin{tabular}[c]{@{}l@{}}GCN-based (BCNet)\end{tabular}                &  IMG  & AM, OO  & SL & \begin{tabular}[c]{@{}l@{}}Objects, Vehicles\end{tabular}  \\ \hline
O, S  & \cite{Yuan_2021_CVPR}                 & \begin{tabular}[c]{@{}l@{}}CompositionalNet (ORM)\end{tabular} &  IMG, MBB & AM, OO & WL & Vehicles \\ \hline
S  &\cite{yuting2021amodal} & CNN-based (Shape Prior)   &  IMG, MM & AM & SL & \begin{tabular}[c]{@{}l@{}}Objects,  Vehicles\end{tabular} \\ \hline
O, S  & \cite{Nguyen_2021_ICCV} & CNN-based (ASBU) &  IMG, MM, OB & AM, OO & WL & \begin{tabular}[c]{@{}l@{}}Objects,  Vehicles\end{tabular} \\ \hline
A, O  &\cite{monnier2021unsupervised} & CNN + MLP (DTI-Sprites) &  IMG & LO, MM, RGB  & UL & Toy dataset   \\ \hline
A, S  &\cite{zhou2021human}  & CNN + GAN  &  IMG, MM     & AM, RGB     & WL & Humans \\ \hline
A, O, S  &  \cite{zheng2021vinv}                   & \begin{tabular}[c]{@{}l@{}}CNN + GAN (CSDNet)\end{tabular} & IMG         & AM, LO, RGB & WL & \begin{tabular}[c]{@{}l@{}}Indoor scenes, \\ Objects, Vehicles\end{tabular} \\ \hline
S  &\cite{Liu_2022_CVPR} & CNN + Geometry-based    &  IMG     & ASL & WL & Road scenes \\ \hline
S  & \cite{sun2022amodal} & Bayesian Generative Model & IMG, MBB & AM & WL & Objects, Vehicles \\ \hline
O  & \cite{Lee_2022_CVPR} & CNN-based (InstaOrderNet)  & IMG, MBB & LO, OO & SL & Objects   \\ \hline
S  & \cite{Reddy_2022_CVPR} & Transformer-based &  IMG & ABB, AM  & SSL & Vehicles \\ \hline
O, S & \begin{tabular}[c]{@{}l@{}} \cite{breitenstein2022amodal} \end{tabular} & CNN-based (amERFNet) & IMG & AM, LO & SL & Vehicles \\ \hline
S  &  \cite{Mohan_2022_CVPR} & CNN-based (APSNet) & IMG & AM & SL & Vehicles \\ \hline
O, S  &\cite{mohan2022perceiving}& CNN-based (PAPS) & IMG & AM, OO & SL & Vehicles\\ \hline
\end{tabular}
\label{tab:main_works}
\end{table*}

\subsection{Amodal Shape Completion}

When considering the shape of an object, amodal completion refers to the use of partial visual evidence to infer the entire shape of the object. Predicting the full shape of the occluded object provides greater insight into the scene, as it helps to infer the actual size and relative depth of objects in the image \citep{kar2015amodal}.

\begin{figure*}[h] 
\centering 
\includegraphics[width=0.8\textwidth]{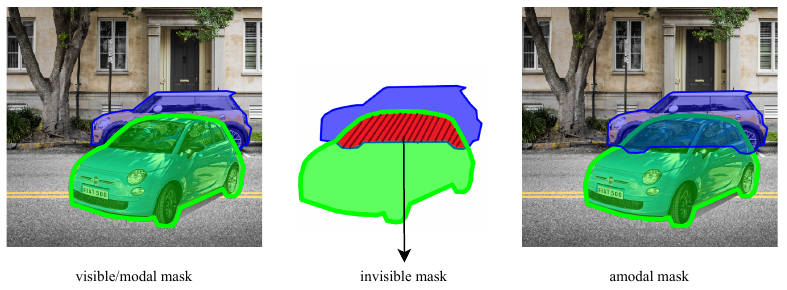} 
\caption{Different types of masks. The Amodel mask of an object is a union of its visible and invisible masks.} 
\label{Fig.masks} 
\end{figure*}

Amodal instance segmentation is the most commonly used technique for predicting the amodal shape of objects. Instance segmentation is used to predict the mask of an object, which can render the shape of the object. Instance segmentation detects instances in an image and assigns a pixel-level label to each instance. This technique is particularly useful in images where objects are very close to, in contact with or overlapping each other. A typical instance segmentation task requires only predicting the visible segmentation mask (also known as the modal mask) of an instance \citep{li2017fully, he2017mask}. However, to understand partially occluded objects in an image, the detector must recognise the full shape of the underlying object, as the absence of the shape can seriously affect object perception. As a result, research on amodal instance segmentation began to emerge.

\begin{figure}[!h] 
\centering 
\scriptsize
\begin{forest}
  for tree={%
    folder,
    grow'=0,
    fit=band,
    align=center,
  }
  [Amodal Shape Completion, rectangle, draw
    [ Instance segmentation with object detection, rectangle, draw
        [ CNN-based methods, rectangle, draw
            [\cite{kar2015amodal}; IBBE \citep{li2016amodal}; \\ASN \citep{qi2019amodal}; BCNet \citep{ke2021deep}, rectangle, draw]
        ]
        [CompositionalNets variants, rectangle, draw
            [Context-aware \citep{wang2020robust}; ORM \citep{Yuan_2021_CVPR}; \cite{sun2022amodal}, rectangle, draw]
        ]
        [Weakly supervised learning, rectangle, draw
            [\cite{Reddy_2022_CVPR}, rectangle, draw]
        ]
    ]
    [Instance segmentation without object detection, rectangle, draw
        [Traditional methods, rectangle, draw
                [\cite{kimia2003euler}; \cite{walton2008improved}; \cite{silberman2014contour};\\  \cite{lin2016computational}; \cite{kihara2016shadows}, rectangle, draw]
        ]
        [CNN-based methods, rectangle, draw
                [AmodalMask \citep{zhu2017semantic}; SeGAN  \citep{ehsani2018segan};\\ 
                ORCNN \citep{follmann2019learning}; 
                \cite{purkait2019seeing};\\
                \cite{hu2019sail};
                Shape Prior \citep{yuting2021amodal}, rectangle, draw]
        ]
        [Weakly supervised learning, rectangle, draw
                [PCNet \citep{zhan2020self}; Amodal-VAE \citep{ling2020variational}; \\
                ASBU \citep{Nguyen_2021_ICCV}, rectangle, draw]
        ]
    ]
    [Other Types of Amodal Shape Representation, rectangle, draw
        [Semantics-aware distance maps, rectangle, draw
            [\cite{zhang2019learning}, rectangle, draw]
        ]
        [Amodal semantic segmentation maps, rectangle, draw
            [amERFNet \citep{breitenstein2022amodal};\\ APSNet \citep{Mohan_2022_CVPR}); 
            PAPS \citep{mohan2022perceiving}, rectangle, draw]
        ]
        [Amodal scene layouts, rectangle, draw
            [MonoLayout \citep{mani2020monolayout}; \cite{narasimhan2020seeing}; \cite{Liu_2022_CVPR}, rectangle, draw]
        ]
    ]
  ]
\end{forest}
\caption{A summary of the methods for amodal shape completion. Amodal instance segmentation is the most commonly used technique for amodal shape completion and can be divided into two types, depending on whether they are combined with object detection techniques. In addition to amodal instance segmentation, we also summarize other types of amodal shape representations.} 
\label{sumshape} 
\end{figure}
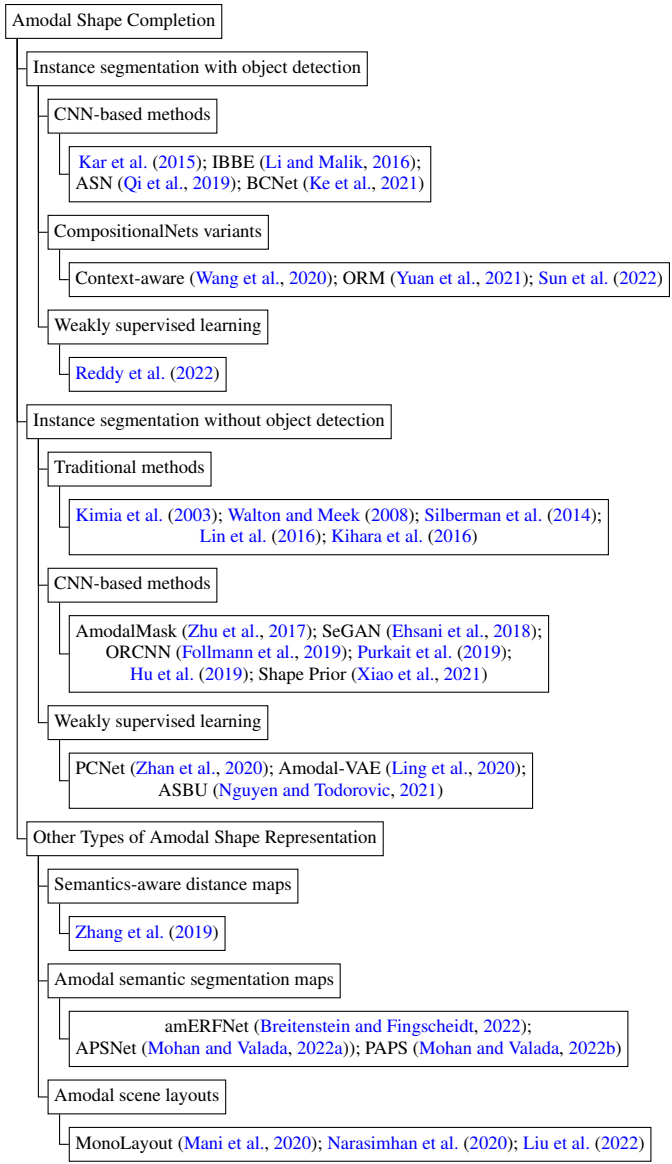

The task of amodal instance segmentation aims to predict the amodal mask of instances, which is the union of the instance‘s visible mask (also known as the modal mask) and the instance‘s invisible mask (see Fig.\ref{Fig.masks}). The amodal masks of partially occluded objects, especially invisible masks, are challenging to predict. When an object is partially hidden by another object of its own class, an algorithm may be confused about where the boundaries of the invisible mask are because features of this category are present in multiple places. Conversely, when an object is partially hidden by another object of a different class, the algorithm may not have enough information about the visible contour of the occluded object to infer the full object mask. Methods for amodal shape completion can be divided into two types, depending on whether they are combined with object detection techniques. One type of approach first identifies the region where the instances are located (usually represented by bounding boxes) by applying object detection techniques, and then performs pixel-level segmentation on the candidate regions. The second type does not rely on object detection but instead directly performs instance segmentation at the pixel level. We discuss each of these two types of approach in the following sections \ref{method_1} and \ref{method_2}. In addition to amodal instance segmentation masks, there are other types of amodal shape representations for objects or scenes, which are discussed in detail in section  \ref{method_3}. Fig.\ref{sumshape} gives a summary of the methods for amodal shape completion.

\subsubsection{Instance Segmentation Methods with Object Detection} 
\label{method_1} 
This type of method mainly adopts a two-stage approach, i.e., detection followed by segmentation. The popular two-stage instance segmentation methods, represented by Mask R-CNN \citep{he2017mask}, first detect each instance of an object class and localize it with a bounding box Region-of-Interest (ROI), and then segment the instance from the background within each ROI to produce an instance mask. These methods are closely connected to object detection works and are usually extended over object detection networks to implement instance segmentation. For example, Mask R-CNN \citep{he2017mask} extends the Faster R-CNN object detection network \citep{ren2015faster} with a Fully Convolutional Network (FCN) \citep{long2015fully} branch to segment objects in the detected bounding boxes, while PANet \citep{liu2018path} further incorporates the multi-layered features of FPN \citep{lin2017feature}, a
feature pyramid network for object detection, to produce better instance segmentation masks. In contrast to these visible instance segmentation methods, amodal shape completion is more concerned with obtaining amodal masks rather than modal masks of instances. Nevertheless, the visible instance segmentation methods have contributed strongly to the development of amodal instance segmentation, as one common way to achieve amodal segmentation is by adding additional amodal modules to the framework designed for visible instance segmentation.

For the first stage of the two-stage approach, in contrast to traditional bounding box detection tasks, which only focus on the visible part of objects, the amodal bounding box detection task extends to the invisible region to predict the full extent of objects in the image. In addition to laying the groundwork for the second stage of instance segmentation, 2D amodal bounding box detection also contributes to the task of 3D amodal detection, such as inferring the 3D bounding box of objects \citep{deng2017amodal}. The information obtained from amodal bounding boxes can also be directly applied in various practical applications. For example, the amodal bounding box detection technique, together with geometric cues, allows the recovery of the real-world size of objects from images \citep{kar2015amodal}. The second stage uses the bounding boxes obtained from the previous stage as ROI features to infer the full extended object mask of the target, including the occluded regions.

\textbf{CNN-based methods:} A small number of studies have focused on the first stage in isolation. One example is a CNN-based framework designed in \citep{kar2015amodal} for predicting the coordinates of amodal bounding boxes only, without object amodal masks. This approach aligns 3D models with object instances in 2D images using keypoint annotations to capture ground-truth amodal bounding boxes, then uses the acquired training data to build amodal bounding box regressors.

Most methods treat the two stages as a continuous process and use amodal bounding box detectors designed to match their framework. One of the first CNN-based amodal instance segmentation methods employing the two-stage approach is presented in IBBE \citep{li2016amodal}. Taking the modal bounding boxes obtained by the CNN-based object detectors \citep{ren2015faster} as input, this method iteratively updates bounding boxes along the direction of higher intensity in the segmentation heatmap and recomputes the heatmap, where the intensity of a heatmap represents the probability of each pixel within the bounding box being an object of interest. By adding additional components to the Mask R-CNN and PANet structures to determine the existence of occlusion in ROI and combining multiple features, ASN \citep{qi2019amodal} extends the design of networks for modal segmentation to amodal segmentation task. Unlike the above approach that treats the input image as a single layer directly, BCNet \citep{ke2021deep}, which combines the Graph Convolutional Network (GCN) \citep{kipf2017semi} with the FCN, models overlapping objects in the same ROI as disjoint bilayers for instance segmentation. This method of leveraging the occlusion relationship between two layers of objects to infer the segmentation result of the occluded object is easier to interpret than single-layer approaches.

\textbf{CompositionalNets variants:} Although CNN-based frameworks are popular choices and provide impressive contributions to two-stage instance segmentation methods, CNN-based predictors sometimes fail to achieve satisfactory performance due to the lack of ability to handle occlusions \citep{WangZXZXY17}. Compared to CNNs, the compositional convolutional neural network (CompositionalNet), which integrates compositional models and deep convolutional neural networks, has shown its robustness to partial occlusion \citep{kortylewski2020compositional}. Designed for classification tasks, CompositionalNet locates the partially occluded objects in an image and then focuses on the visible part of the object to perform classification. According to \citep{wang2020robust}, however, the predictions of CompositionalNet can be severely disrupted by the surrounding context when there is heavy object occlusion due to the lack of explicit separation between the object representation and the context.

To address the failure of CompositionalNet under heavy occlusion, a context-aware CompositionalNet is introduced in \citep{wang2020robust} to disentangle the object representation and the context. By extending CompositionalNet with an additional corner-based voting mechanism, the model is able to estimate amodal bounding boxes for even heavily occluded objects robustly. Similar to \citep{kar2015amodal}, this method only predicts the amodal bounding boxes but not the amodal masks. In contrast, \citep{Yuan_2021_CVPR} proposes a CompositionalNet-based model trained with amodal bounding box supervision to perform amodal instance segmentation. As well as following the idea of \citep{wang2020robust} to separate the context from the objects, this method extends the generative model of CompositionalNet with an Occlusion Reasoning Module (ORM). The ORM is used to detect inconsistent segmentations and correct them by estimating the occlusion order of nearby objects. The corrected segmentations are then passed back into the network to self-correct and improve the prediction results. Also inspired by CompositionalNet, a Bayesian generative model of the neural network features is used to replace the fully-connected classifier in the CNN to infer the amodal segmentation in \citep{sun2022amodal}. The Bayesian model then uses the probability distribution to explain the features of the image, including the object classes and amodal segmentation. However, this method requires extensive shape priors and is therefore limited to rigid objects such as vehicles.

\textbf{Weakly supervised learning:} While most previous work using deep learning has adopted supervised learning, this approach typically requires large amounts of dense pixel-level instance segmentation annotations. Manually adding amodal segmentation annotations to large datasets is often highly time-consuming, e.g., an estimated 15 minutes per image for the BSDS dataset and 18 minutes per image for the COCO dataset \citep{zhu2017semantic}. On the other hand, auto-generated data not only provides an affordable and precise alternative for amodal annotations, but also offers the possibility of automatic occlusion supervision. For example, a clip-art-based approach \citep{Reddy_2022_CVPR} treats unoccluded objects that are automatically mined over time in time-lapse images as longitudinal self-supervision. By using time-lapse images from fixed cameras, different unoccluded objects captured over time are composited back into the same scene to produce images containing occlusion and corresponding amodal segmentation ground truth. These composite images are then passed into a transformer-based network \citep{liu2021swin} to produce amodal bounding boxes and amodal instance segmentation masks, which use the auto-generated amodal segmentation masks as supervision. However, this approach requires huge numbers of images and long training times for each scene, and the ability to generalise to new scenes is not yet clear. More efficient methods for amodal instance segmentation are expected in the future.

\subsubsection{Instance Segmentation Methods without Object Detection}
\label{method_2}

This type of method aims to predict amodal instance masks directly without first obtaining the object's amodal bounding box.

\textbf{Traditional methods:} Early amodal shape completion work in this area typically follows Gestalt principles \citep{kimia2003euler} and assumes the most likely shape or curve for the invisible region. For example, Euler Spirals \citep{kimia2003euler, walton2008improved}, straight lines and parabolas \citep{silberman2014contour}, and cubic Bézier curves \citep{lin2016computational} have been applied to the domain of amodal shape completion. Synthetic images with partially overlapping objects provide the shape prior to Shape Boltzmann machine modelling in \citep{kihara2016shadows}. However, models trained on synthetic image datasets consisting of simple binary shapes are not likely to produce
the desired results when recovering the full mask of object instances. Although shape or curve-based assumptions are feasible in toy examples with simple shapes, it is extremely difficult to make assumptions that apply to different real-world objects in highly cluttered natural scenes.

\textbf{CNN-based methods:} With the rise of deep learning, recent studies have attempted to use deep learning models for shape completion due to their advantages in automatically learning features for different occlusion patterns. A straightforward idea is to use existing models designed for modal segmentation and train them on amodally annotated data. However, despite being able to predict the amodal mask behind occlusion, the performance of such models decreases sharply with increasing occlusion, as they are designed for segmenting the visible part of objects only \citep{zhu2017semantic}.

In order to specialise in amodal segmentation, a deep neural network called AmodalMask is proposed in \citep{zhu2017semantic} for predicting the amodal masks from image patches. It adopts the SharpMask \citep{pinheiro2016learning} architecture and trains on the amodal ground truth. This ground truth is provided by human annotators marking the amodal masks of objects by imagining their shapes behind occlusions. A pioneering work, \citep{zhu2017semantic} includes an extensive analysis showing that semantic amodal segmentation is a well-posed annotation task as different human annotators tend to agree on the amodal completion content of the invisible parts. Unlike AmodalMask, which provides class-agnostic predictions, the predictions of Occlusion R-CNN (ORCNN) \citep{follmann2019learning} are class-specific. This algorithm adds additional segmentation branches over Mask R-CNN to make both amodal and modal mask predictions. It also simultaneously provides the invisible mask by computing the difference between the amodal and modal masks. However, both AmodalMask and ORCNN learn a mapping relationship from the content of the entire visible area to the amodal mask, which means that different occlusions may affect the prediction of the amodal mask for even the same occluded object. To mitigate the misleading effect of occlusion features, \citep{yuting2021amodal} concentrates on the visible region of the occluded object and introduces shape prior knowledge to refine and post-process the coarse amodal mask predicted by Mask R-CNN. The category-specific shape prior is modelled using an autoencoder to extract the embeddings of ground-truth amodal masks and then compute their cluster centre.

Some methods are designed based on computer-generated images and precise annotations rather than using manually annotated amodal masks. Applying rendering techniques to create computer-generated scenes is an economical and fast way of acquiring data, and solving the problem of manually annotating the invisible parts of objects, which is expensive and subject to the bias of human annotators. In order to make the resulting image closer to reality, some studies build sophisticated and realistic scenes to obtain photo-realistic images. Snapshots taken in 3D synthetic indoor scenes are used in \citep{ehsani2018segan,purkait2019seeing} to train a CNN network for amodal segmentation. Screenshots extracted from the GTA-V video game are used to train the MaskJoint model, which jointly infers the amodal and modal object masks \citep{hu2019sail}. Nevertheless, models trained with rendered images may suffer from domain gaps and fail to generalise to real scenes.

\textbf{Weakly supervised learning:} Aside from recent advances in supervised learning methods, efforts have also been made to explore ways to relieve supervision demand for amodal mask completion. A self-supervised method for training the amodal segmentation mask network is introduced in \citep{zhan2020self}. With the idea of partial completion, this method treats once amodal mask segmentation as equivalent to multiple partial completion processes. Partial completion means that only the portion of the target object that is occluded by the given occluder will be completed. Training of the proposed partial completion network is achieved by artificially placing an occluder over the ground-truth modal mask, so that the learned network can recover the amodal mask with the same pattern. The effectiveness of the network relies on correctly predicting the occlusion order between objects, so errors in the occlusion ordering graph of each instance can seriously affect the output of the network. As an improvement, ASBU \citep{Nguyen_2021_ICCV} uses the occlusion boundary as a replacement for the occluder's modal mask, freeing the prediction of amodal masks from the constraints of occlusion ordering graphs. This weakly-supervised design requires only ground-truth modal masks to train a model to predict amodal masks, circumventing the problem of the lack of ground-truth amodal masks. Similarly, a variational autoencoder (VAE) designed to complete amodal masks without ground-truth amodal segmentation annotations is presented in \citep{ling2020variational}. The advantage of this probabilistic model is that it allows for ambiguity and thus offers multiple solutions rather than generating a unique amodal mask. However, in this work, the variability learned by the model is only evaluated in driving scenes with a limited number of categories of rigid objects. Future work would be needed to demonstrate the feasibility of the VAE approach over a wider range of categories and non-rigid objects.

\subsubsection{Other Types of Amodal Shape Representation}
\label{method_3}

While most studies consider amodal masks of instances as mentioned above, some studies use other forms to represent amodal shapes.

\textbf{Semantics-aware distance maps}, which describe the visibility of different areas of an object, can also be used to represent an amodal segmentation result. A two-stage CNN-based architecture is proposed in \citep{zhang2019learning} to provide semantics-aware distance maps for all objects in a single image, where amodal bounding box proposals are utilised in an ROI-Align layer \citep{he2017mask} to capture instance-level and global-level feature maps for each instance.

\textbf{Amodal semantic segmentation maps} reveal information about the background in addition to considering the objects in the image. Following the idea of grouping different objects in \citep{purkait2019seeing}, such as static and traffic objects, pixels of different semantic classes in an image are grouped for the semantic segmentation task \citep{breitenstein2022amodal}. Also, one channel is used to indicate whether the pixel is visible or occluded. While amodal semantic segmentation provides more background information, it does not distinguish between instances of the same semantic class. As a result, amodal semantic segmentation does not provide occlusion relations between different instances of the same semantic class. To simultaneously produce the pixel-level semantic segmentation map and the instance-level amodal instance segmentation mask, an amodal panoptic segmentation network, APSNet, is proposed in \citep{Mohan_2022_CVPR}. By dividing the components in an image into the "stuff" class (amorphous or uncountable) and the "thing" class (countable), the amodal panoptic segmentation task attempts to predict both pixel-level semantic segmentation maps for the "stuff" class, and amodal instance segmentation masks for the "thing" class. To achieve simultaneous predictions, the APSNet employs parallel semantic and amodal instance segmentation heads prior to final fusion. However, for the "stuff" class, the APSNet only provides the traditional semantic segmentation map of the visible part, not the amodal semantic segmentation map with the insights of the invisible part. In addition, APSNet may fail in the case of heavy occlusion, as the prediction relies on the features from the visible area. As a countermeasure against heavy occlusion, an amodal mask refiner that embeds non-occluded objects' features to complement the amodal features is proposed in PAPS \citep{mohan2022perceiving}. This amodal mask refiner can be incorporated into other amodal segmentation methods, such as ASN \citep{qi2019amodal} and APSNet \citep{Mohan_2022_CVPR}.

\textbf{Amodal scene layouts} provide insight into the layout of scenes that are occluded or truncated in image space. This task allows the perception of the road or indoor layouts from images, greatly facilitating practical applications such as self-driving or path planning \citep{narasimhan2020seeing, Liu_2022_CVPR}. For road scenes, some studies aim to predict the complete bird's eye view (BEV) layout even if the image is partially occluded. The MonoLayout \citep{mani2020monolayout} is designed to generate the amodal scene layout in BEV containing road areas, footpaths and vehicles from the input image, even if the scene is partially covered. It is an adversarial learning network containing two decoders and discriminators, one for static elements such as roads and the other for dynamic elements such as vehicles. Compared to the previous approaches, which require pixel-level semantic labels as supervision, a weaker supervised approach is proposed in \citep{Liu_2022_CVPR}. The proposed approach converts layout predictions into parametric attributes to interact with human annotators, avoiding costly manual pixel-by-pixel labelling. However, most amodal scene layout research is focused on road scenes, more research will be needed to expand to various scenarios, such as indoor scene layouts \citep{narasimhan2020seeing}.

\subsection{Amodal Appearance Completion}

Amodal appearance completion aims to recreate the appearance of invisible regions of object instances or backgrounds. General image inpainting methods \citep{yu2019free,xie2019image} restore a user-selected missing area in an image to create a visually-reasonable output, without concern for which object(s) the missing region belongs to. In contrast, amodal appearance completion algorithms autonomously identify the partially-occluded objects and their hidden regions that need to be reconstructed. The amodal appearance completion task typically involves breaking the scene down into a series of individual objects and then recovering the entire RGB appearance of those target objects that are partially occluded.

Most of the current amodal appearance completion methods rely on amodal masks inferred from amodal shape completion. They treat the amodal shape completion and appearance completion as a comprehensive pipeline. The shape completion process provides information about the invisible or/and visible regions of the object, which serve as the input to the appearance completion process. The appearance completion process then fills in the missing pixels for the invisible areas of the object based on the input signal.

\begin{figure}[!h] 
\centering 
\scriptsize
\begin{forest}
  for tree={%
    folder,
    grow'=0,
    fit=band,
    align=center,
  }
  [Amodal Appearance Completion, rectangle, draw
    [ Object-centric Representations for Toy Datasets, rectangle, draw
        [ MONet \citep{burgess2019monet}; IODINE \citep{greff2019multi}; \\GENESIS \citep{engelcke2020genesis}; Slot Attention  \citep{locatello2020object}; \\DTI-Sprites \citep{monnier2021unsupervised}, rectangle, draw]
    ]
    [Category-specific Studies, rectangle, draw
        [Vehicles \citep{yan2019visualizing}, rectangle, draw]
        [Humans \citep{zhou2021human}, rectangle, draw]
        [Food: PizzaGAN \citep{Papadopoulos_2019_CVPR}, rectangle, draw]
    ]
    [Methods for Complex Scenes, rectangle, draw
        [Frameworks based on synthetic data, rectangle, draw
            [SeGAN \citep{ehsani2018segan}; \cite{dhamo2019object}, rectangle, draw]
        ]
        [Schemes for real images, rectangle, draw
            [PCNet \citep{zhan2020self}; CSDNet \citep{zheng2021vinv}, rectangle, draw]
        ]
    ]
  ]
\end{forest}
\caption{A summary of the methods for amodal appearance completion. These methods can be divided into three groups based on the application scenario: object-centric representations for toy datasets, methods designed for specific classes of objects, and methods for more complex general scenarios.} 
\label{sumappear} 
\end{figure}
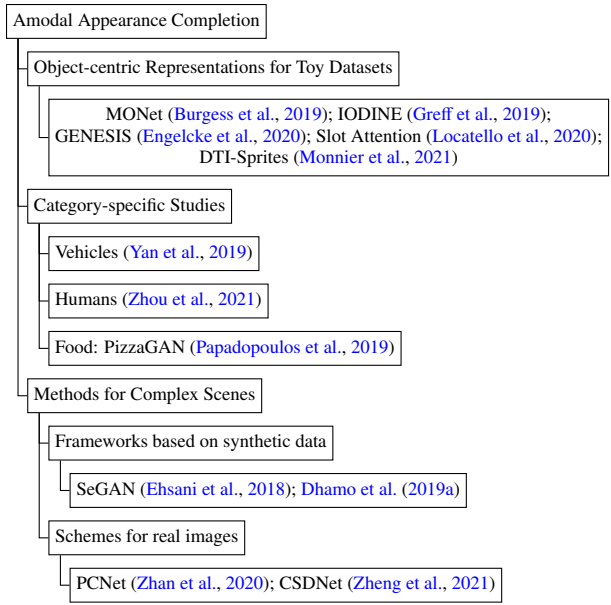

Based on the application scenario, amodal appearance completion methods can be broadly divided into three groups: object-centric representations for toy datasets, methods designed for specific classes of objects, and methods for more complex general scenarios. Fig.\ref{sumappear} gives a summary of the methods for amodal appearance completion.

\subsubsection{Object-centric Representations for Toy Datasets}
One way to approach the amodal appearance completion task is to decompose images into object-centric representations. The object-centric representation assumes that an image is composed of a mixture of latent components, such as multiple objects and regions. Most models treat objects as 2D or 2.5D (with depth) sprites and learn object-centric representations by reconstructing the input image. MONet \citep{burgess2019monet} combines a VAE and a recurrent attention network to obtain objects and reconstruct their amodal appearance. Also building on the VAE framework, IODINE \citep{greff2019multi} uses iterative variational inference to jointly infer both object appearance and object segmentation. Inspired by these two works, GENESIS \citep{engelcke2020genesis} introduces an autoregressive prior over the object-centric latent variables to help capture the dependencies between scene components. The autoregressive prior is learned with a sequential inference, which sequentially infers later object variables conditioned on others. Using iterated attention, Slot Attention \citep{locatello2020object} simplifies the multiple encode-decode steps in MONet, IODINE and GENESIS with a single encoding step to improve computational efficiency.

However, these object-centred approaches have only been shown to be effective on simple toy examples, where the objects have very distinct colours compared to the background. As an improvement, DTI-Sprites \citep{monnier2021unsupervised} presents a framework that jointly learns the object prototypes and occlusion/transformation predictors to reconstruct images, and applies this framework to real images. This method first decomposes an image into multiple object prototypes, then uses a greedy method to combine the prototypes and find the combination that is most similar to the original image. Nevertheless, the results are heavily influenced by the surrounding context when reconstructing everyday scenes.

All of these methods are trained in an unsupervised manner to learn representations of different objects in the image, and then use the object representations to reconstruct the original image. Although this approach has been shown to work in toy examples and real images with simple backgrounds, it has not yet been successfully applied to complex real scenes.

\subsubsection{Category-specific Studies}

Some works have focused on completing the amodal appearance of specific classes of occluded objects.

\textbf{Vehicles:} An iterative framework is proposed to complete the appearance of the occluded parts of vehicles in \citep{yan2019visualizing}. After getting the amodal mask of a vehicle from the segmentation completion module, the amodal mask is passed through a GAN-based two-path network to recover the appearance of invisible regions of the vehicle. One path is trained to fill in the colour of the occluded area, and the other path is trained to inpaint the entire vehicle according to the image context. These two paths share the same network parameters and only the first path is used during testing. Finally, the results are refined by multiple iterations in the segmentation completion module and the appearance recovery module. As we know, vehicles are a relatively straightforward object class, as there is not much intra-class variation in their amodal shape and RGB appearance compared to other types of objects.

\textbf{Humans:} Compared to vehicles, humans are more complex in amodal shape because they can assume different poses and have greater variation in appearance. Similar to the previous approach for completing vehicle appearance, \citep{zhou2021human} proposes a two-stage framework for amodal appearance completion of occluded people in images, where the first stage segments the human amodal mask and the second stage recovers the appearance within the obtained amodal mask. This approach uses an Unet \citep{ronneberger2015u} with partial convolution \citep{liu2018image} as the appearance completion network and is equipped with a Parsing Guided Attention (PGA) module. The PGA module contains two attention streams, one for distinguishing different body parts and the other for establishing relationships between visible and invisible areas.

\textbf{Food:} The idea of amodal appearance completion has also been applied to teach machines to understand food recipes. A generative model is used in PizzaGAN \citep{Papadopoulos_2019_CVPR} to remove pizza toppings from images by filling in the appearance of the detected parts that are occluded by the to-be-removed ingredients. The ingredients are predicted by classifying the object classes appearing in a given image. Nevertheless, these methods designed for specific objects have limited applications, and what works for one type of object may not perform equally well on others.

\subsubsection{Methods for Complex Scenes}

\textbf{Frameworks based on synthetic data:} An obstacle to the development of models for amodal appearance completion in more complex scenes is a lack of suitable training data. As there are no large-scale datasets with ground-truth amodal appearances for all objects, many studies have utilised computer-generated synthetic datasets. Considering amodal segmentation and appearance completion of occluded objects as an integral work, SeGAN \citep{ehsani2018segan} employs a cGAN for the amodal appearance completion step to generate the appearance of the occluded part of the object. Since the amodal appearance completion step takes only one input, the amodal mask predicted from the previous segmentation step, the appearance completion leans heavily on the predictions of the segmentation and ignores contextual information. Another approach trained on synthetic data, \citep{dhamo2019object} uses an object completion branch to infer a fully visible RGBA-D representation of each object from the input RGB image, the predicted amodal instance masks and object category predictions. However, when applied to real images, accurate estimation of pixel-level depth values from a single image can also be a challenge.

Both of these methods rely on amodal ground-truth annotations to train the model, rather than automatically understanding the whole scene. Such supervised learning frameworks are therefore no longer applicable when the lack of amodal appearance ground-truth annotations. It is also not clear how well approaches trained on synthetic images can generalise to real-world images, given the lack of ground-truth data for testing.

\textbf{Schemes for real images:} A method to complete the occluded appearance without amodal annotations as supervision is explored in PCNet \citep{zhan2020self}. This approach uses a self-supervised framework on top of the concept of partial completion, which progressively handles order recovery, amodal masks and appearance completion. The concept of partial completion involves two principles. Firstly, in the presence of multiple occluded areas, the occluded objects can be progressively completed by processing only one occluded area at a time. Secondly, the network can be trained to partially complete an occluded object by deliberately pruning the object and then having the network recover that just-pruned part. Similarly, the framework first completes the amodal mask with one network, and then feeds both the visible and invisible masks of the object into another network to produce the appearance of the invisible part of that object. However, the authors could not quantitatively assess the effectiveness of the amodal appearance completion method on real images due to the absence of appearance completion ground truth.

In order to evaluate the generalisation performance of the amodal appearance completion method in different scenes as well as rendered ones, CSDNet \citep{zheng2021vinv} proposes the use of pseudo-ground-truth generated on real images. CSDNet is built on the idea that the fully visible objects detected on each layer of an image are actually occluders for the next layer. PICNet \citep{zheng2019pluralistic}, a probabilistic framework with two parallel GANs, is adopted here to generate repaired appearance. The fully visible objects are therefore extracted and the network fills in the appearance of these occluded areas for the next layer. The reconstructed image is passed back to the network, which again detects and segments the fully visible objects and fills in the appearance for the next layer. For real images, this approach uses pseudo-ground-truth generated from a trained synthesis model as weakly-supervised labels and fine-tunes the synthetically-trained CSDNet. However, the pseudo-ground-truth they produce has visible differences from real images. How to generate plausible and realistic pseudo-ground-truth for real images needs further exploration.

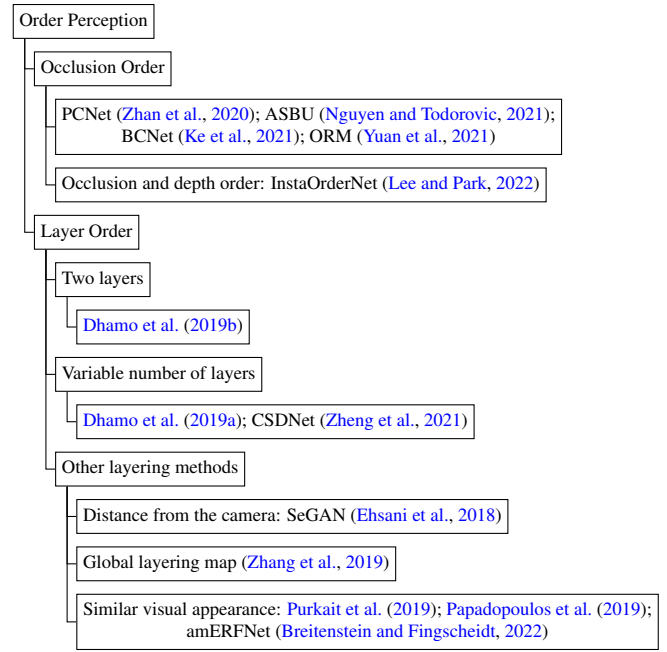
\begin{figure}[] 
\centering 
\scriptsize
\begin{forest}
  for tree={%
    folder,
    grow'=0,
    fit=band,
    align=center,
  }
  [Order Perception, rectangle, draw
    [ Occlusion Order, rectangle, draw
        [PCNet \citep{zhan2020self}; ASBU \citep{Nguyen_2021_ICCV};\\ BCNet \citep{ke2021deep}; ORM \citep{Yuan_2021_CVPR}, rectangle, draw]
        [Occlusion and depth order: InstaOrderNet \citep{Lee_2022_CVPR}, rectangle, draw]
    ]
    [Layer Order, rectangle, draw
        [Two layers, rectangle, draw
            [\cite{dhamo2019peeking}, rectangle, draw]
        ]
        [Variable number of layers, rectangle, draw
            [\cite{dhamo2019object}; CSDNet \citep{zheng2021vinv}, rectangle, draw]
        ]
        [Other layering methods, rectangle, draw
            [Distance from the camera: SeGAN \citep{ehsani2018segan}, rectangle, draw]
            [Global layering map \citep{zhang2019learning}, rectangle, draw]
            [Similar visual appearance: \cite{purkait2019seeing}; \cite{Papadopoulos_2019_CVPR};\\ amERFNet \citep{breitenstein2022amodal}, rectangle, draw]
        ]
    ]
  ]
\end{forest}
\caption{A summary of the approaches to perceptual order in image amodal tasks. Two common ways to describe the order relationship between objects in a scene are occlusion order and layer order.} 
\label{ordersum} 
\end{figure}

\subsection{Order Perception}

\begin{figure*}[h] 
\centering 
\includegraphics[width=0.8\textwidth]{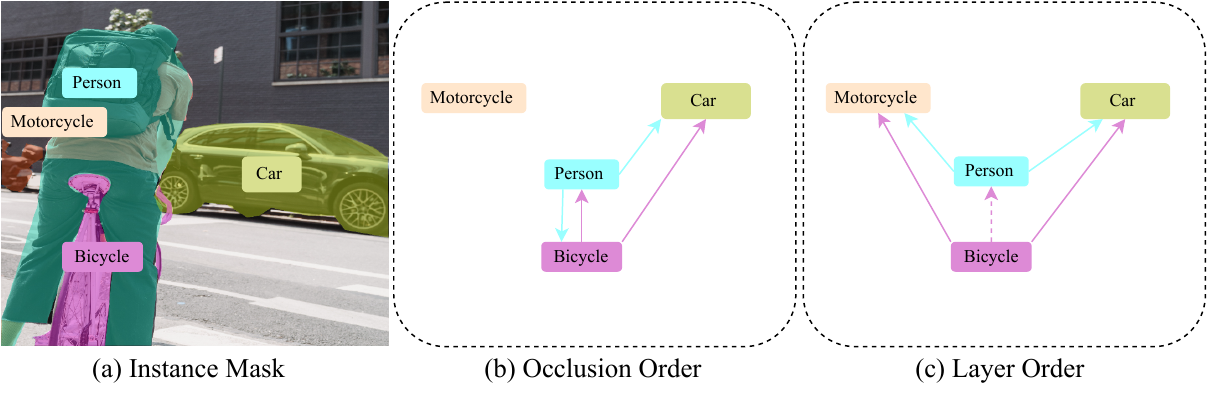} 
\caption{Two expressions of order perception. (a) Example image with modal instance masks. (b) illustrates the occlusion order by using a directed graph with arrows pointing from the occluder instance to the occludee instance. (c) shows a layer order with a depth-based strategy, where the arrow points to a more distant object and the dashed line indicate the two instances share a partial overlap in depths.} 
\label{Fig.main2} 
\end{figure*}

For the amodal completion, a question that naturally arises is what is the correct order between overlapping objects. In particular, some amodal shape/appearance algorithms rely on the inferred order to complete their tasks. A common way to infer the order is to learn common occlusion/depth patterns and their effect on the order between objects. For example, learners can work out that horse riders and horses usually occlude each other and share the same depth. Once equipped with order perception, a vision system can use the inferred order between objects to assist in completing the shape and appearance of invisible areas in the image. Fig.\ref{ordersum} gives a summary of the approaches to perceptual order in image amodal tasks. Generally, there are two ways to describe the order relationships between objects in a scene, namely occlusion order and layer order (see Fig.\ref{Fig.main2}). The occlusion order considers the occlusion relationship between objects, involving occluder (the object which occludes others) and occludee (the object being occluded). Order relationships can include no occlusion or bi-directional occlusion (two objects mutually occlude each other). On the other hand, layer order considers the depth relationship between objects, usually determined by the distance from the object to the camera. The two ordering schemes are complementary to each other. For example, objects on different layers may not directly occlude each other.

\subsubsection{Occlusion Order}

For amodal completion tasks, some methods do not explicitly determine the occlusion relationship between objects, instead caring only about whether or not there is occlusion within the ROI \citep{qi2019amodal, follmann2019learning}. However, these methods may fail to deal with the segmentation conflicts between heavily overlapping objects due to the lack of order reasoning between the occludee and the occluder.

Compared to approaches that only consider a single occluded object within the ROI, BCNet \citep{ke2021deep} separates the occludee and the occluder of an ROI into two layers, where the shape of the occluder is detected using GCN and the occludee is inferred based on the occluder. Occlusion boundaries can be defined via repeated regions between layers, providing important information for decoupling occlusion relationships. Similarly, ORM \citep{Yuan_2021_CVPR} also assumes that there is only one occluder and one occludee in the overlapping region of the two bounding boxes, and reassigns each pixel in the overlapping region to one of the two objects based on a designed pixel-level competition. The object with more pixels in the overlapping area is defined as the occluder. The occlusion order graph of the whole image is then obtained by comparing the two objects in each overlapping region. With the occlusion order, the network is able to improve the amodal segmentation by iterative self-correction.

Explicitly inferring the occlusion order between objects not only makes the final segmentation result easier to interpret, but is also crucial for amodal completion methods that require accurate occlusion information. For example, PCNet \citep{zhan2020self} requires the union of all occluders as a priori knowledge to complete the occluded object, including those that do not directly occlude the target object. Therefore obtaining the occlusion order of all objects in the scene is a requirement for PCNet. This method also starts with an inference of the occlusion order between two neighbouring objects. For two instances where the modal masks are adjacent to each other, the trained mask completion network is used to predict the amodal masks for each instance. The difference between the modal and amodal masks of both instances are then compared, and the instance with the greater difference is regarded as the occludee. Similarly, ASBU \citep{Nguyen_2021_ICCV} also assumes that, in a given pair of adjacent instances, the instance with the larger invisible region is the one being occluded. Given the pairwise occlusion order for all adjacent object pairs, the occlusion order of objects in the whole scene can be deduced.

Occlusion order is not the only way to explain the interrelationship between objects in a cluttered scene. At the same time, considering depth alone is not sufficient to determine whether there is an occlusion relationship between objects. Therefore, recent work has attempted to jointly predict occlusion order and depth order. For instance, InstaOrderNet \citep{Lee_2022_CVPR} replaces the last layer of the CNN network with two fully connected layers to produce both occlusion order and depth order. In addition to simultaneously predicting the occlusion order and the depth order, an alternative approach is to consider the interactions between objects together with their position in the scene, such as the layer order we discuss in the following section.

\subsubsection{Layer Order}

One way to completely represent where objects are located within the scene and how they relate to each other is to build a 3D model of the entire scene. However, this is often an excessively complex option due to the high computational and memory consumption associated with generating a complete 3D model. Directly extrapolating the depth map from a single image, on the other hand, is insufficient to capture the geometry and appearance information for invisible parts of occluded objects. A trade-off solution is to use a layered structure to represent the original image, which is simpler than a 3D model but able to capture the required information. Different work may use different assumptions about the layers of an image, such as a two-layer structure that divides the image into foreground and background, or a more complex multi-layer structure. After designing a specific assumption about the layers for the algorithm,  we can produce the layer order of the objects. The layer order is then used, along with other features of the visible parts of the objects, to predict the shape and appearance of the partially occluded regions.

Scene parsing methods have been proposed to produce the layer order for visible regions in an image \citep{zhang2018hierarchical}. In contrast, amodal completion focuses not only on the visible regions of the image but also learns hidden information about the invisible regions. The idea of using a layered structure to represent the image is also widely used in the task of novel view synthesis \citep{tucker2020single, amberbir2020generative, luvizon2021adaptive}. Similar to amodal completion, the task of novel view synthesis also requires predicting what lies behind the visible object. The difference is that, depending on the angle of the novel view relative to the original view, it may not be necessary to predict much-hidden information. For small shifts in view, usually only a few pixels near the edges of the visible objects need to be predicted.

\textbf{Two layers:} The simplest layering structure divides the scene into just two layers; for example, a foreground layer containing all object instances and a background layer without objects. \citep{dhamo2019peeking} uses this representation with a fully–convolutional CNN to predict a segmentation mask for the foreground layer. All foreground pixels are then discarded from the image, leaving the background layer to be filled in using a GAN-based approach. Due to the two-layer representation, the subsequent background completion task will not be fooled by the appearance of objects in the foreground layer. However, using a predetermined number of layers to represent scenes of varying complexity limits both the flexibility and performance of this type of method.

\textbf{Variable number of layers:} In contrast, some methods employ a variable number of layers that are adapted to the scene. A natural idea is to decompose the image into a series of object-containing layers plus an object-free background layer. A model that determines the number of layers based on the number of detected object instances is presented in \citep{dhamo2019object}. This approach uses a re-composition block (consisting of three convolutions followed by ReLUs) for layer ordering, using the predicted instance depth and the mask as priors. However, this object-wise layering method can lead to the scene being divided into too many layers when it contains a substantial number of objects. A variation on this idea is to share layers among particular objects so that the number of layers required can be reduced. The fully visible objects in each layer are used to decompose the scene in CSDNet \citep{zheng2021vinv}. In each layer, the network determines and extracts the instances that are classified as fully visible. The missing areas of the image are then filled with appearance and then passed back to the network. The completed image is again subjected to the step of extracting fully visible objects, thus progressively obtaining the layer order of the scene.

\textbf{Other layering methods:} In addition, the distance of the object from the camera can be incorporated into the layering representation. For example, pairwise occlusion order can be leveraged to assess the distance of objects from the camera, as the occluder in a pair of objects is usually closer to the camera than the occludee \citep{ehsani2018segan}. Other ideas for the layering order include using a global layering map that indicates the pixel-level visibility level of each object in the image \citep{zhang2019learning}, or treating objects of similar visual appearance as the same layer \citep{purkait2019seeing, Papadopoulos_2019_CVPR, breitenstein2022amodal}.

\section{DATASETS and EVALUATION}
\begin{figure*}[h] 
\centering 
\includegraphics[width=0.8\textwidth]{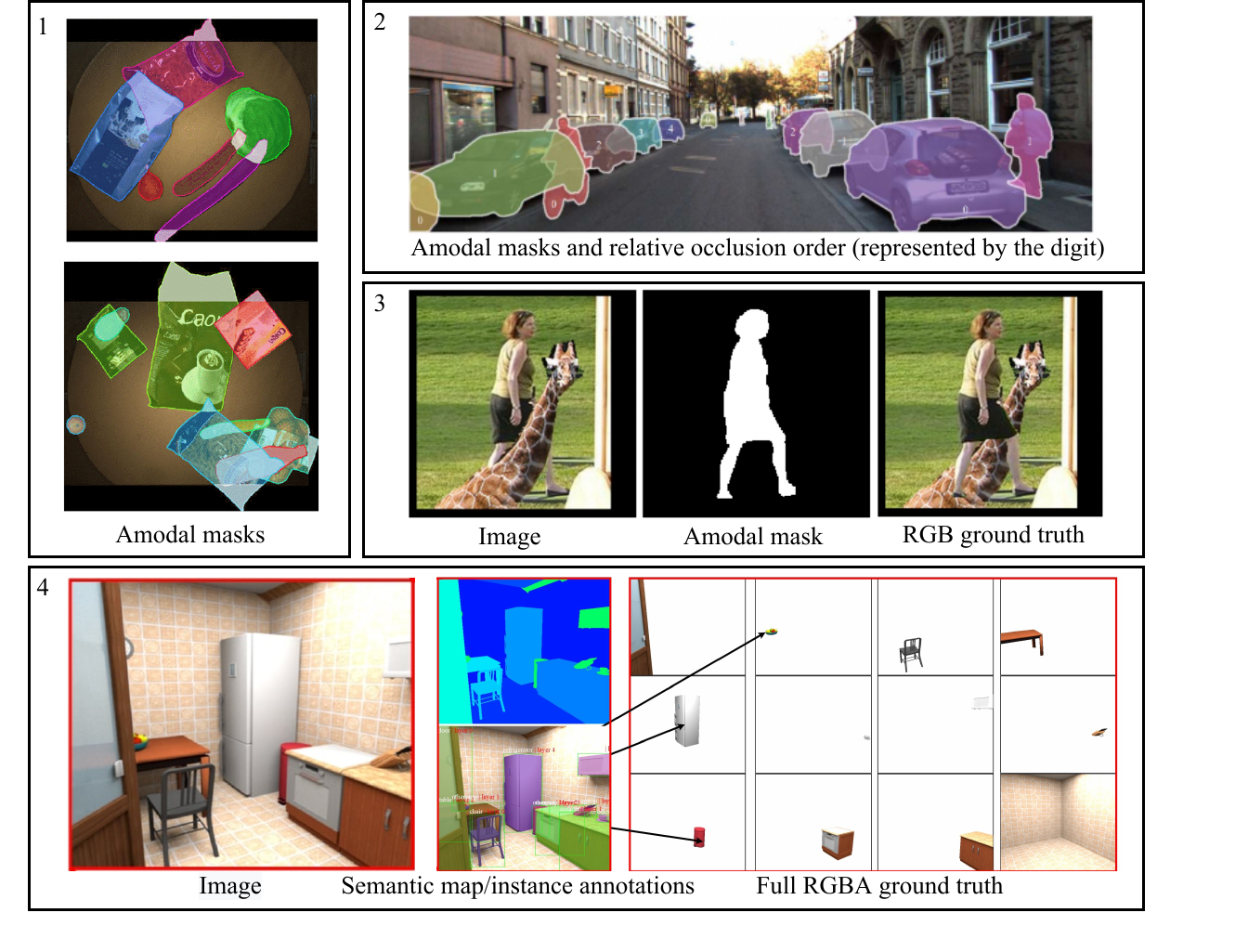} 
\caption{Examples of amodal datasets with ground truth annotations. 1) D2S Amodal \citep{follmann2019learning}, 2) KINS \citep{qi2019amodal}, 3) AHP \citep{zhou2021human}, 4) CSD \citep{zheng2021vinv}.} 
\label{Fig.datasetexample} 
\end{figure*}

Datasets play a key role in the development of the field of image amodal completion. High-quality datasets tend to improve the quality of the trained models and the accuracy of predictions. Yet there are relatively few datasets available for this emerging field. Although the image amodal completion task in general can be broken down into three subtasks and each subtask serves a different goal, the subtasks are sometimes inextricably intertwined. For example, amodal appearance completion often requires the support of amodal masks and order perception. Therefore, relevant datasets may be designed to support multiple subtasks at the same time. Fig.\ref{Fig.datasetexample} gives examples of amodal datasets with ground truth annotations for supporting image amodal completion tasks.

In this section, we first introduce the popular datasets designed for image amodal completion and their main data collection methods. We then review the common evaluation metrics adopted for each subtask and provide the latest benchmark results on the relevant amodal datasets. Finally, we discuss the out-of-distribution generalization of existing methods.

\subsection{Data Collection}
Existing related datasets adopt three main approaches for data collection, namely manual annotation, clip-art-based image composition and the generation of 2D images from 3D synthetic scenes.

\subsubsection{Manual Annotation}
With the aim of supporting the development of machine vision systems with human-like amodal perception capabilities, a number of studies have involved human annotation of datasets. Human observers are asked to annotate images with their amodal perception, i.e., to label the full extent region of each object rather than just the visible pixels. Despite the high cost of acquiring annotations, the survey in \citep{zhu2017semantic} shows very strong agreement between human annotators in predicting occluded regions, so the use of manual annotations tends to produce consistent amodal shape results. Nevertheless, current datasets produced through manual annotation only provide amodal instance segmentation masks and/or object order, but not the appearance of invisible regions.

\subsubsection{Clip-art-based Image Composition}
Clip-art-based image compositing refers to cutting patches from one image and superimposing them as layers onto the target image to simulate occlusion. Compared to manual annotation, clip-art-based image compositing is a low-cost way of collecting data and can additionally provide the appearance of occluded areas alongside the amodal instance masks and object order. However, this method can only provide the amodal appearance data for occlusions produced by added objects; if a complex scene is used as the background, any pre-existing occlusions cannot be resolved. Additionally, there may be inconsistencies or artefacts produced by the superimposition process that can affect the realism and quality of the resulting dataset.

\subsubsection{Generated Images from Synthetic Scenes}
Using 2D images generated from 3D synthetic scenes allows for more flexible and detailed annotation. One way of obtaining amodal annotations is to align the target instance to an approximate 3D model \citep{kar2015amodal}. Another common way is to capture the ground truth of each instance or frame directly from a computer-rendered 3D scene. A dataset produced using synthetic 3D scenes has the highest degrees of freedom for annotation and can robustly provide amodal instance masks, full appearance, occlusion order and layer order for all objects in the scene. Nevertheless, data from synthetic 3D scenes are usually limited in terms of diversity and realism.

\subsection{Datasets}

The availability of public amodal datasets allows different frameworks and algorithms to be trained, tuned and compared with each other. High-quality public datasets are therefore of great value. In this sub-section, we classify popular public datasets for image amodal completion tasks according to their applications and summarise the key information about them in Table \ref{tab:datasets}.

\begin{table*}[!h]
\renewcommand\arraystretch{1.2}
\footnotesize
\centering
\caption{Summary of publicly available amodal datasets. \# represents the number of the following items. pix* represents common image sizes in datasets where the image size is not fixed. 100,395* refers to the dynamic objects on the training set. $-$ represents not available. 36 Mil* refers to the unoccluded instances. OI: occluded instances. Avg. OR \%: Avg. occlusion rate \%, which means the average IoU between two objects. AM: whether the dataset annotation provides amodal masks for instances. AA: whether the dataset annotation provides RGB amodal appearance for instances. Order: whether the dataset annotation provides order relations between objects. R/S: Real or Synthetic. App.: the application scenarios of these datasets. Objects refer to common objects in a natural context.}
\begin{tabular}{|l|l|l|l|l|l|l|l|l|l|l|l|}
\hline
Dataset   & \#Images & Resolution & \#Classes & \#Instances & \#OI      & \begin{tabular}[c]{@{}l@{}}Avg. \\OR \%\end{tabular} & AM & AA & Order & R/S &   APP.  \\ \hline
COCOA \citep{zhu2017semantic}     & 5,073    & 275K pix*   & $-$         & 46,314       & 28,106     & 18.8       & $\checkmark$  & $-$  & $\checkmark$     & R        & Objects        \\ \hline
COCOA-cls \citep{follmann2019learning} & 3,499    & 275K pix*   & 80        & 10,562       & 5,175      & 10.7       & $\checkmark$  & $-$  & $\checkmark$     & R     &Objects           \\ \hline
InstaOrder \citep{Lee_2022_CVPR}       & 100,623   & 275K pix*    & 80        & 503,939     &  $-$   & $-$       & $-$ & $-$  & $\checkmark$     & R       & Objects    \\ \hline
D2S Amodal \citep{follmann2019learning}     & 5,600    & 1400×1920  & 60        & 28,720       & 16,337     & 15         & $\checkmark$  & $-$  & $   -      $     & S &Objects\\ \hline
SAIL-VOS \citep{hu2019sail}  & 111,654   & 800×1280   & 162       & 1,896,296   & 1,653,980 & 56.3       & $\checkmark$  & $\checkmark$  & $\checkmark$     & S     &Objects     \\ \hline
KINS \citep{qi2019amodal}      & 14,991   & 1242×375   & 8         & 190,626     & 99,964    & 19.8       & $\checkmark$  & $-$  & $\checkmark$     & R      &Vehicle          \\ \hline
OVD \citep{yan2019visualizing}       & 34,100   & 500×375*    & 196       & –            &–           & –           & $\checkmark$  & $\checkmark$  & $    - $     & S   &Vehicle   \\ \hline
WALT \citep{Reddy_2022_CVPR}       & 15 Mil & 4K/1080p    & 2   & 36 Mil*  & $-$ & $-$      & $\checkmark$ & $\checkmark$  & $\checkmark$     & R+S &Vehicle  \\ \hline \begin{tabular}[c]{@{}l@{}}
Amodal Cityscapes \\ \citep{breitenstein2022amodal}\end{tabular}      & 3,475   & 1024x2048    & 19       & 100,395*     & $-$    & 0.06       & $\checkmark$  & $-$  & $\checkmark$      & S      &Vehicle      \\ \hline
\begin{tabular}[c]{@{}l@{}}KITTI-360-APS \\\citep{Mohan_2022_CVPR} \end{tabular}       & 61,168   & 1408x376    & 17      & 215,342      & $-$     & $-$        & $\checkmark$ & $-$  & $-$     & R    &Vehicle        \\ \hline
\begin{tabular}[c]{@{}l@{}}BDD100K-APS \\ \citep{Mohan_2022_CVPR} \end{tabular}      & 202   & 1280 × 720    & 16      & 49,058      & $-$     & $-$        & $\checkmark$ & $-$  & $-$     & R      &Vehicle      \\ \hline
AHP \citep{zhou2021human}       & 56,599   & non-fixed  & 1         &  56,599      &  –       & –           & $\checkmark$  & $\checkmark$  & $ -       $     & R+S   & Human   \\ \hline
DYCE \citep{ehsani2018segan}      & 5,500    & 1000×1000  & 79        & 85,975       & 70,766     & 27.7       & $\checkmark$  & $\checkmark$  & $   -      $     & S   &Indoor        \\ \hline
OLMD \citep{dhamo2019object}      & 13,000   & 384×512    & 40        & –           & –         & $-$          & $\checkmark$  & $\checkmark$  & $\checkmark$     & S          &Indoor   \\ \hline
CSD \citep{zheng2021vinv}       & 11,434   & 512×512    & 40        & 129,336     & 74,596    & 26.3       & $\checkmark$  & $\checkmark$  & $\checkmark$     & S     &Indoor        \\ \hline
\end{tabular}
\label{tab:datasets}
\end{table*}

\subsubsection{Common Objects Datasets}

The COCOA dataset \citep{zhu2017semantic} provides an amodal segmentation for each of the five thousand images collected from COCO \citep{lin2014microsoft}. The amodal segmentation includes amodal and modal segmentation masks for each annotated segment and pairwise depth-ordering between segments. The annotation of each image was manually performed by one human annotator. Rather than using the common COCO categories \citep{lin2014microsoft}, COCOA \citep{zhu2017semantic} provides labels for spatial extents such as grass and water, as well as specific objects such as people and flowers. As a result, there is a wide range of labels in the dataset, but each label has a limited number of instances. As opposed to the class-agnostic COCOA dataset, the COCOA-cls dataset \citep{follmann2019learning}, which builds on both COCOA and COCO, offers class-specific amodal annotation for objects in the COCO categories.

The InstaOrder dataset \citep{Lee_2022_CVPR} is also created from COCO \citep{lin2014microsoft} but with additional manual annotation of the geometric ordering of the instances in the images. The annotations include both the occlusion order among objects and the depth order based on the distance from the object to the camera. In contrast to the COCOA \citep{zhu2017semantic} and COCOA-cls \citep{follmann2019learning} datasets, the InstaOrder dataset \citep{Lee_2022_CVPR} further provides bidirectional occlusion order and depth order of overlap types, but only contains modal segmentation masks instead of amodal segmentation masks.

The D2S Amodal dataset \citep{follmann2019learning} consists of images of groceries and everyday products in a top-down view, and it is constructed by amodally annotating all images in the Densely Segmented Supermarket (D2S) dataset \citep{follmann2018mvtec}. Occlusions in the dataset are artificially generated by overlaying multiple cropped individual instances, so that amodal and modal segmentation masks for the occluded instances can be easily provided. The annotation also includes a category label for each object.

The SAIL-VOS Dataset \citep{hu2019sail} contains video frames of both indoor and outdoor scenes from the GTA-V Environmental Simulator. To obtain amodal segmentation masks for partially occluded objects, the visibility of all objects in a single frame is toggled one by one to capture each individual instance. In addition, category labels are provided manually for each object, including 48 of the COCO categories and 114 other common object categories. Depth information is also available, but it is calculated at the pixel level rather than at the instance level.

\subsubsection{Vehicle Datasets}

The KITTI INStance (KINS) dataset \citep{qi2019amodal} provides extra amodal instance segmentation masks and relative occlusion order for representative categories of objects in KITTI images \citep{geiger2012we}. The objects in the representative categories include the general categories of people and vehicles, where people are further divided into three sub-categories such as 'pedestrians' and 'cyclists', and vehicles are further divided into five sub-categories such as 'cars' and 'trams'. The additional annotations are manually annotated by three annotators and disagreements are resolved using crowd-sourcing. As the dataset consists of real images captured in driving scenarios, it does not provide the full RGB appearance of the occluded objects.

The Occluded Vehicle Dataset (OVD) \citep{yan2019visualizing} generates images of occluded vehicles and their annotations by manually superimposing other objects over the vehicles in the images collected from the Cars dataset \citep{krause20133d}. In addition to the pre-existing vehicle categories, amodal segmentation masks and  the fully visible RGB appearance of vehicles are provided by manually labelling the unoccluded vehicles in the original images. The Deep Harmonization technique \citep{tsai2017deep} is used to construct the composite images. The objects superimposed on the car images are extracted from the COCO \citep{lin2014microsoft} and CityScape \citep{cordts2016cityscapes} dataset.

The Watch And learn Time-Lapse (WALT) Dataset \citep{Reddy_2022_CVPR} encompasses time-lapse images of urban traffic scenes from six cameras set up by the authors and six public live streams for one year. To further provide a dataset for amodal detection and segmentation studies, the authors generated the Clip Art WALT Dataset (CWALT) from WALT Dataset, which clips unoccluded objects in the same scene from time-lapse images captured by stationary cameras and superimposes them onto the generated scene. Furthermore, two additional datasets are generated from the WALT dataset for evaluation: the Rendered WALT Dataset, which is generated from the rendering of a parking lot 3D scene, and the Stationary Objects WALT Dataset, which is generated by mining unoccluded stationary objects.

The Amodal Cityscapes Dataset \citep{breitenstein2022amodal} is an amodal semantic segmentation dataset generated from the Cityscapes Dataset \citep{cordts2016cityscapes}. Occlusion is simulated by clipping instances from other images and superimposing them onto the target image. For each generated image, a corresponding ground truth amodal semantic segmentation map is provided, including the semantic class of the pixel and whether the pixel is occluded.

The KITTI-360-APS dataset and the BDD100K-APS dataset are both large-scale urban scenes with amodal panoptic annotations \citep{Mohan_2022_CVPR}. The KITTI-360-APS dataset is an extension of the KITTI-360 dataset \citep{liao2022kitti} and contains ten "stuff" classes and seven "thing" classes; the BDD100K-APS dataset is developed from the Berkeley Deep Drive (BDD100K) dataset \citep{yu2018bdd100k} and contains ten "stuff" classes and six "thing" classes. The annotation performs in a semi-automatic manner, i.e. after generating pseudo-amodal instance masks or semantic segmentation maps using a pre-trained model, the human annotators then manually correct these generated pseudo-annotations. The model generating the pseudo-annotations is then fine-tuned. This process is then iterated until the whole dataset is annotated. As a result, the annotation includes visible semantic segmentation maps for the "stuff" classes and amodal instance segmentation masks for the "thing" classes.

\subsubsection{Human Datasets}
The Amodal Human Perception (AHP) dataset \citep{zhou2021human} consists of 56,599 images collected from multiple large-scale instance segmentation and detection datasets, including COCO \citep{lin2014microsoft}, VOC \citep{everingham2010pascal} (with semantic boundaries \citep{hariharan2011semantic}), LIP \citep{gong2017look}, Objects365 \citep{shao2019objects365} and Open Images V4 \citep{kuznetsova2020open}. Each image includes an unoccluded instance of a person that is fully visible within the image, along with its fine pixel-level segmentation mask. The dataset spans a wide range of scene categories, including outdoor environments such as ballparks and streets, and indoor environments such as stages and homes. Only the annotation of a single human instance is provided for each image, even if there is more than one person in the image, and the annotated instance is usually standing. Other instances are pasted onto the unoccluded human instances to form additional artificial occlusion cases. Thus for each composite image, it is possible to obtain the amodal/modal mask and the RGB appearance of the human instance.

\subsubsection{Indoor Datasets}
The DYCE dataset \citep{ehsani2018segan} contains 5.5k snapshots captured from 11 synthetic 3D indoor scenes. The camera is set at roughly human height and intentionally avoids areas without objects, taking 500 different views of each scene. The scenes include 5 living rooms and 6 kitchens, each containing an average of 60 objects. Each image is provided with two global instance segmentation masks, one for visible regions and the other for invisible regions in the image. As the objects in the 3d scene are rendered independently, the dataset also provides the full RGB appearance of each object as well as the background.

The OLMD dataset \citep{dhamo2019object} captures the complete RGBA-D representation of objects and layout layers from the same image, using synthetic scenes from the SunCG dataset \citep{song2017semantic} and Stanford 2D-3D dataset \citep{armeni2017joint}. In addition, the dataset extracts depth maps and categories for each instance in the scene. The captured objects and layout layers can be further represented as Layered Depth Images (LDI) \citep{gortler1998layered} based on the depth map. Although this rendered dataset provides images with various object-wise layers for each scene, the images are somewhat low resolution and unrealistic.

The CSD dataset \citep{zheng2021vinv} is built upon the 3D CAD models of SUNCG \citep{song2017semantic}. It contains a total of 11.4k images rendered from more than 2k indoor environments using the Maya rendering software, in which the camera viewpoints are randomly sampled. All images include more than five object instances and each instance and the empty environment is rendered separately. Each image is accompanied by a global-level semantic graph, directed graphs of layer order and pairwise occlusion order as well as instance-level dense annotations (category, bounding box, modal and amodal mask, full RGB appearance). In addition to all foreground objects, the full RGB appearance is also provided for the occlusion-free background.

These well-annotated datasets have facilitated the development of the field of image amodal completion, and novel evaluation metrics for amodal completion have been developed in response to multiple types of annotation for different tasks.

\subsection{Evaluation Metrics and Latest Benchmark Results on the Amodal Datasets}
\label{metrics}

Different evaluation metrics are used for the three subtasks of image amodal completion. A framework for image amodal completion may involve only one subtask or several subtasks at the same time; therefore, which subtasks need to be evaluated should be determined by the intention of the framework. The appropriate amodal dataset for evaluation should also be chosen based on the subtasks being addressed. In this sub-section, we present the common evaluation metrics for each subtask of image amodal completion and provide the latest benchmark results on the relevant amodal datasets.

\subsubsection{Amodal Shape Completion}

Most amodal datasets designed for amodal shape completion tasks focus on instance masks, while some consider other forms of amodal shape representation, including datasets for amodal semantic segmentation and amodal panoptic segmentation. Compared to amodal instance segmentation, amodal semantic segmentation provides a broad understanding of the image's content by classifying its pixels into different semantic categories instead of individual objects. Amodal panoptic segmentation takes amodal semantic segmentation one step further by differentiating instances of the same semantic class for countable objects (i.e. thing classes). Due to the different objectives, different evaluation metrics are also used for different types of segmentation masks.

\textbf{Amodal instance segmentation:}
For the amodal shape completion task, in which the goal is to predict the amodal segmentation mask for the instances, most researchers employ the commonly-used metrics for object detection and modal instance segmentation tasks, including IoU (Intersection over Union), MIoU (Mean Intersection over Union), Precision, Recall, F1-score, mAP (mean Average Precision), AP Across Scales and AR (Average Recall). Among them, mAP is usually the most popular metric, which is sometimes referred to as AP in evaluations. As shown in Table \ref{ais}, some commonly used amodal instance segmentation datasets include the COCOA \citep{zhu2017semantic, follmann2019learning} and KINS \citep{qi2019amodal}.

\begin{table}[]
\renewcommand\arraystretch{1.2}
\centering
\footnotesize
\caption{Results on the COCOA \citep{zhu2017semantic, follmann2019learning} and KINS \citep{qi2019amodal} with methods designed for amodal instance segmentation. The results are taken from the paper \citep{zhan2020self, ke2021deep, zheng2021vinv, yuting2021amodal, Reddy_2022_CVPR}. All scores are in \%.}
\begin{tabular}{|l|ll|}
\hline
\multirow{2}{*}{Method}                  & \multicolumn{1}{l|}{COCOA} & KINS\;\; \\ \cline{2-3} 
                                         & \multicolumn{2}{l|}{mAP}       \\ \hline
ORCNN \citep{follmann2019learning}       & \multicolumn{1}{l|}{33.2}  & 29.0 \\ \hline
ASN (Mask-RCNN)  \citep{qi2019amodal}    & \multicolumn{1}{l|}{34.0}  & 31.1 \\ \hline
PCNet \citep{zhan2020self}               & \multicolumn{1}{l|}{30.3}  & 29.3 \\ \hline
BCNet \citep{ke2021deep}                 & \multicolumn{1}{l|}{32.7}  & 28.8 \\ \hline
CSDNet (Mask-RCNN) \citep{zheng2021vinv} & \multicolumn{1}{l|}{34.1}  & 31.5 \\ \hline
Shape Prior \citep{yuting2021amodal}     & \multicolumn{1}{l|}{35.4}  & 32.0 \\ \hline
\cite{Reddy_2022_CVPR}                   & \multicolumn{1}{l|}{33.1}  & 27.9 \\ \hline
\end{tabular}
\label{ais}
\end{table}

\begin{table}[]
\renewcommand\arraystretch{1.2}
\centering
\footnotesize
\caption{Results on the Amodal Cityscapes dataset \citep{breitenstein2022amodal} with method designed for amodal semantic segmentation. The results are taken from the paper \citep{breitenstein2022amodal}. K is the number of groups of semantic classes in the dataset. $MIoU^{vis}$, $MIoU^{inv}$, and $MIoU^{total}$ represent the MIoU of the visible region, the MIoU of the invisible region, and the MIoU of the total region (including both visible and invisible regions) in the image, respectively. All scores are in \%.}
\begin{tabular}{|l|l|l|l|l|}
\hline
Method    & K & $MIoU^{vis}$ & $MIoU^{inv}$ & $MIoU^{total}$ \\ \hline
\multirow{2}{3cm}{amERFNet \citep{breitenstein2022amodal}}
 & 3 & 61.9     & 18.5       & 58.0         \\ \cline{2-5}
 & 4 & 62.7     & 23.6       & 59.1         \\ \hline
\end{tabular}
\label{ass}
\end{table}

\begin{table}[!]
\renewcommand\arraystretch{1.2}
\centering
\caption{Results on the KITTI-360-APS and BDD100K-APS datasets \citep{Mohan_2022_CVPR} with methods designed for amodal panoptic segmentation. The results are taken from the paper \citep{mohan2022perceiving}. All scores are in \%.}
\footnotesize
\begin{tabular}{|l|ll|ll|}
\hline
\multirow{2}{*}{Method}                                                    & \multicolumn{2}{l|}{KITTI-360-APS} & \multicolumn{2}{l|}{BDD100K-APS} \\ \cline{2-5} 
 & \multicolumn{1}{l|}{APQ\;\;}   & APC   & \multicolumn{1}{l|}{APQ\;\;}  & APC  \\ \hline
\begin{tabular}[c]{@{}l@{}}APSNet\\ \citep{Mohan_2022_CVPR}\end{tabular}   & \multicolumn{1}{l|}{42.9}  & 59.0  & \multicolumn{1}{l|}{46.3} & 47.3 \\ \hline
\begin{tabular}[c]{@{}l@{}}PAPS\\ \citep{mohan2022perceiving}\end{tabular} & \multicolumn{1}{l|}{44.6}  & 61.4  & \multicolumn{1}{l|}{48.7} & 50.4 \\ \hline
\end{tabular}
\label{aps}
\end{table}

In addition to standard segmentation metrics, the difference between the amodal mask and the modal mask is used to evaluate the extent to which an object is occluded. The area ratio \citep{li2016amodal} is the ratio of the area of the intersection of the modal and amodal masks to the area of the amodal mask, so that the ratio of non-occluded objects is close to 1 and the ratio of heavily occluded objects is close to 0. The concept of area ratio has sparked interest among researchers to consider the model's effectiveness across varying levels of occlusion. As a result, a number of studies have evaluated the segmentation performance separately at different occlusion levels \citep{wang2020robust, Yuan_2021_CVPR}.

Apart from quantitative measurements, qualitative results presented through the visualisation of prediction examples are also widely used as some datasets do not provide the ground truth annotations for evaluating amodal instance segmentation. In this case, the performance of an amodal shape completion model can be measured by the satisfaction of human observers \citep{ling2020variational}.

\textbf{Amodal semantic segmentation:} Some evaluation metrics for amodal semantic segmentation are based on MIoU. The amERFNet \citep{breitenstein2022amodal} proposes to calculate the MIoU for the visible regions, the invisible regions and the total regions (including both visible and invisible regions) of the image separately to indicate the quality of the amodal semantic segmentation. As shown in Table \ref{ass}, the main dataset used for the amodal semantic segmentation experiments is the Amodal Cityscapes \citep{breitenstein2022amodal}.

\textbf{Amodal panoptic segmentation:}
 Amodal panoptic quality (APQ) and amodal parsing coverage (APC) are common metrics for amodal panoptic segmentation. APQ takes extra account of invisible areas compared to the standard panoptic quality metric \citep{kirillov2019panoptic}, and APC takes account of the size of the instances compared to APQ \citep{Mohan_2022_CVPR}. As shown in Table \ref{aps}, the major datasets used for the amodal panoptic segmentation experiments are the KITTI-360-APS and BDD100K-APS \citep{Mohan_2022_CVPR}.

\subsubsection{Amodal Appearance Completion}
For the amodal appearance completion task, studies adopt a number of different metrics to measure the performance of the recovered images. Most commonly-used metrics focus on pixel differences between prediction and ground truth, including MSE (Mean Square Error), RMSE (Root Mean Square Error), and PSNR (Peak Signal-to-noise Ratio). Alternatively, Distance metrics, such as L1 and L2 distances \citep{ yan2019visualizing}, can be used to measure the per-pixel error. Absolute pixel differences do not necessarily map to human perceptual differences, so some metrics try to consider more perceptual differences. For example, SSIM (Structural Similarity Index) takes luminance, contrast and structure into account \citep{dhamo2019object}. Particularly, for the appearance recovered using GANs, the similarity of the generated results to the ground truth can be measured using FID (Fréchet inception distance), which is the Fréchet distance between two multivariate Gaussians (generated and ground-truth images) \citep{heusel2017gans}. As with amodal shape completion, human studies using qualitative model results are also used to determine which method produces a better complete object appearance, according to human observers \citep{ehsani2018segan}. As shown in Table \ref{apc}, some commonly used amodal appearance completion datasets include the CSD \citep{zheng2021vinv} and AHP \citep{zhou2021human}.

\subsubsection{Order Perception}

For order perception, the design of the framework determines whether the occlusion order or the layering order should be evaluated.

\textbf{Occlusion order:}
For occluded pairwise instances, the average accuracy of pairwise depth ordering is a common metric. This is the mean ratio of the number of correct predictions to the number of all occluded pairwise instances. Likewise, precision, recall and F1 scores can be used to assess the occlusion order of instance pairs by considering the ratio of correctly predicted occlusion orders. The average precision of pairwise depth ordering, on the other hand, is the mean ratio of the number of correct predictions to the number of all predicted pairwise orderings. As shown in Table \ref{oo}, the common amodal datasets used for predicting occlusion order are the COCOA \citep{zhu2017semantic}, KINS \citep{qi2019amodal} and InstaOrder \citep{Lee_2022_CVPR}. In addition to considering the depth ordering of all occluded instance pairs, it is also possible to consider only pairwise instances that have a certain degree of occlusion \citep{Yuan_2021_CVPR}. As an extension, the average accuracy or average precision can be evaluated over various thresholds.

\begin{table}[]
\renewcommand\arraystretch{1.2}
\centering
\caption{Results on the CSD \citep{zheng2021vinv} and AHP \citep{zhou2021human} with methods designed for amodal shape completion. The results are taken from the paper \citep{zheng2021vinv, zhou2021human}. S. and R. represent synthesized and real validation images. $-$ represents no results reported.}
\footnotesize
\begin{tabular}{|l|lll|ll|}
\hline
\multirow{3}{*}{Method} & \multicolumn{3}{l|}{CSD} & \multicolumn{2}{l|}{AHP} \\ \cline{2-6} 
 & \multicolumn{1}{l|}{\multirow{2}{*}{RMSE}} & \multicolumn{1}{l|}{\multirow{2}{*}{SSIM}} & \multirow{2}{*}{PSNR}  & \multicolumn{1}{l|}{L1 (S.)} & FID (S.) \\ \cline{5-6} 
 & \multicolumn{1}{l|}{}                      & \multicolumn{1}{l|}{}                      &                        & \multicolumn{1}{l|}{L1 (R.)} & FID (R.) \\ \hline
\multirow{2}{*}{\begin{tabular}[c]{@{}l@{}}SeGAN\\ \citep{ehsani2018segan}\end{tabular}} & \multicolumn{1}{l|}{\multirow{2}{*}{0.12}} & \multicolumn{1}{l|}{\multirow{2}{*}{0.81}} & \multirow{2}{*}{21.42} & \multicolumn{1}{l|}{0.11}    & 23.01    \\ \cline{5-6} 
 & \multicolumn{1}{l|}{}                      & \multicolumn{1}{l|}{}                      &                        & \multicolumn{1}{l|}{0.10}    & 35.21    \\ \hline
 \cite{dhamo2019object}                                                                                      & \multicolumn{1}{l|}{0.15}                  & \multicolumn{1}{l|}{0.76}                  & 20.32                  & \multicolumn{1}{l|}{$-$}     & $-$      \\ \hline
\multirow{2}{*}{\cite{yan2019visualizing}}                                                                    & \multicolumn{1}{l|}{\multirow{2}{*}{$-$}}  & \multicolumn{1}{l|}{\multirow{2}{*}{$-$}}  & \multirow{2}{*}{$-$}   & \multicolumn{1}{l|}{0.09}    & 27.15    \\ \cline{5-6} 
 & \multicolumn{1}{l|}{}                      & \multicolumn{1}{l|}{}                      &                        & \multicolumn{1}{l|}{0.09}    & 36.23    \\ \hline
\multirow{2}{*}{\begin{tabular}[c]{@{}l@{}}PCNet\\ \citep{zhan2020self}\end{tabular}}                    & \multicolumn{1}{l|}{\multirow{2}{*}{0.11}} & \multicolumn{1}{l|}{\multirow{2}{*}{0.82}} & \multirow{2}{*}{23.16} & \multicolumn{1}{l|}{0.09}    & 18.5     \\ \cline{5-6} 
 & \multicolumn{1}{l|}{}                      & \multicolumn{1}{l|}{}                      &                        & \multicolumn{1}{l|}{0.09}    & 28.3     \\ \hline
\begin{tabular}[c]{@{}l@{}}CSDNet\\ \citep{zheng2021vinv}\end{tabular}                                    & \multicolumn{1}{l|}{0.06}                  & \multicolumn{1}{l|}{0.91}                  & 35.24                  & \multicolumn{1}{l|}{$-$}     & $-$      \\ \hline
\multirow{2}{*}{\cite{zhou2021human} }                                                                     & \multicolumn{1}{l|}{\multirow{2}{*}{$-$}}  & \multicolumn{1}{l|}{\multirow{2}{*}{$-$}}  & \multirow{2}{*}{$-$}   & \multicolumn{1}{l|}{0.05}    & 13.85    \\ \cline{5-6} 
 & \multicolumn{1}{l|}{}                      & \multicolumn{1}{l|}{}                      & & \multicolumn{1}{l|}{0.06}    & 19.49    \\ \hline
\end{tabular}
\label{apc}
\end{table}

\begin{table}[]
\renewcommand\arraystretch{1.2}
\centering
\caption{Results on the COCOA \citep{zhu2017semantic}, KINS \citep{qi2019amodal} and InstaOrder \citep{Lee_2022_CVPR} datasets with methods designed for predicting occlusion order. The results are taken from the paper \citep{Nguyen_2021_ICCV, Lee_2022_CVPR}. Prec. is the pair-wise precision and Acc. is the pair-wise accuracy for ordering recovery. $-$ represents no results reported. All scores are in \%.}
\footnotesize
\begin{tabular}{|l|l|l|l|l|l|}
\hline
Method                                                                & Dataset       & Recall & Prec. & F1 & Acc. \\ \hline
\multirow{3}{*}{\begin{tabular}[c]{@{}l@{}}PCNet-M\\ \citep{zhan2020self}\end{tabular}} & COCOA &    82.3    &84.5&82.8&87.1\\ \cline{2-6} 
& KINS          &94.6&91.6&92.5&92.5\\ \cline{2-6} 
& InstaOrder &59.1&76.4&63.0&$-$\\ \hline
\multirow{3}{*}{\begin{tabular}[c]{@{}l@{}}$InstaOrderNet^{o}$\\ \citep{Lee_2022_CVPR} \end{tabular}} & COCOA          &88.6&85.3&86.1&$-$\\ \cline{2-6}
& KINS          &98.7&94.5&96.0&$-$\\ \cline{2-6} 
& InstaOrder &89.3&79.8&80.6&$-$\\ \hline

\multirow{2}{2.3cm}{ASBU \citep{Nguyen_2021_ICCV}} & COCOA          &$-$&$-$&$-$&90.3\\ \cline{2-6}
& KINS          &$-$&$-$&$-$&92.6\\ \hline
\end{tabular}
\label{oo}
\end{table}

\textbf{Layer order:} For layer order, different layering designs lead to varied evaluation metrics. One method for evaluation is with respect to depth error, with metrics such as RMSE (root mean square error), REL (Relative Error) and accuracy with threshold (\% of $d_i$ such that $max(\frac{d_i}{d_i^*},\frac{d_i^*}{d_i}) = \delta < thr$, where $d_i$ is the actual depth and $d_i^*$ is the estimated depth). These metrics are calculated based on a comparison of the estimated depth with the actual depth. Accuracy metrics are typically reported based on three different thresholds ($thr=1.25, 1.25^2, 1.25^3$) in the literature \citep{Lee_2022_CVPR}. For the layer order with unknown actual depths, distance metrics can be used to quantify the ordering accuracy. For example, the Damerow-Levenstein distance is used in \citep{Papadopoulos_2019_CVPR} to calculate the minimum number of operations required to change the ground-truth layer order into the predicted one. Another idea of evaluation is to convert the layer order for each pair of overlapping instances into a pairwise occlusion relationship \citep{zheng2021vinv}. In addition, each layer can be evaluated individually to measure the performance of the model at a particular layer \citep{dhamo2019object}.

\subsection{Out-of-Distribution Generalization}
The evaluation of out-of-distribution (OOD) generalization is crucial for amodal completion tasks as it assesses the robustness and reliability of a model beyond the distribution of the training data. In particular, many existing amodal methods are trained on synthetic datasets, and the domain gap between real and synthetic images highlights the need to evaluate their performance on the OOD examples. To evaluate the performance of the methods on OOD examples, one approach is to test with unseen datasets \citep{Reddy_2022_CVPR, Lee_2022_CVPR}. Alternatively, models can be trained on non-occluded images and evaluated on partially occluded ones \citep{sun2022amodal}. In the case of amodal appearance completion, the scarcity of real datasets with accurate texture ground truth for occluded areas makes it necessary to resort to closer-to-natural images for evaluating OOD generalization. This includes using 3D CAD models to project near-realistic appearance images \citep{ehsani2018segan}, manually collecting and annotating real-world images \citep{yan2019visualizing}, or manually editing visually reasonable images \citep{zhou2021human}.

\section{APPLICATIONS}
Image amodal completion is an important part of many real-world applications. Including automatic generation of amodal annotations, scene editing and restructuring, Diminished Reality, robotic gripping systems, self-driving and novel view synthesis.

\subsection{Automatic Generation of Amodal Annotations}
Amodal annotations are more demanding on human annotators than modal annotations, as annotation of invisible parts requires good drawing skills from the annotator. Thus, obtaining amodal annotations manually is extremely expensive and time-consuming. Luckily, a large number of existing datasets are available with modal annotations. With the method of amodal shape completion, it is possible to leverage existing image segmentation datasets by converting modal annotations into pseudo-amodal annotations. Well-trained models are able to provide amodal masks that contain more detail and are more natural than manually-annotated masks. Existing user studies have shown that model-generated amodal masks are already comparable to human-labelled amodal masks in a dataset containing simple categories of objects \citep{zhan2020self}, and that human observers even prefer the automatically-generated masks \citep{ling2020variational}.

\subsection{Scene Editing and Restructuring}
The amodal completion task makes it possible to decompose a scene into separate complete objects and backgrounds, while also providing ordering between objects. This makes it possible to edit and manipulate the scene, for example, by swapping the order between objects, removing objects, copying objects or repositioning objects \citep{zhan2020self}. These features provide users with a convenient way to edit their photos, especially to remove large occluders naturally. For example, removing people or cars from holiday photos, or removing rubbish from real estate photos. It provides a more realistic result when regenerating the image, as the in-depth understanding of the scene includes even invisible areas. This application of interactive scene editing offers a new dimension to image editing. Amodal completion is also particularly useful for privacy protection applications. For sensitive objects, the amodal completion method allows the removal of objects while naturally filling in the invisible parts of the background, which provides a more realistic and aesthetically pleasing result than traditional mosaic methods.

\subsection{Diminished Reality}
DR (Diminished Reality) is a complementary technology to AR (Augmented reality). AR allows one to enhance or add virtual content to their perception of the physical world. DR allows people to remove undesired entities from real scenes captured through a camera, thus gaining a more realistic perspective. For example, in an AR furniture shopping application, people use AR technology to place virtual chairs, tables, and decorations into a space for selection. However, if the shot is not taken in an ideal empty room, the original objects, such as the existing furniture, can get in the way of the consumer's experience and perception. This is where the use of DR technology helps to remove unnecessary clutter and make room for new furniture that the consumer wants to try. DR does not simply remove obstacles, but also automatically fills in missing image information according to the features of the floor or wall \citep{gkitsas2021panodr, pintore2022instant}. Image amodal completion is the perfect way to achieve DR technology.

\subsection{Robotic Gripping Systems} Recent research has demonstrated the utility of image amodal completion in robotic gripping systems \citep{wada2019joint, inagaki2019detecting, wada2018instance}. Robotic gripping systems are commonly used in warehouses where there are many items, and the items are often partially occluded by other items. Failures in gripping tasks usually occur when the robot is unable to locate a heavily occluded target object. To find an occluded object, the robot system needs to have amodal completion ability, i.e., it must be able to understand the shape of the occluded object and the occlusion relationships between objects. This enables the robot to plan to pick sequences and remove obstacles to reach the target. Even when grasping fully visible objects, the amodal completion ability helps the robot system to avoid collisions with other objects that could lead to grasping failure.

\subsection{Self-driving}
Self-driving vehicles require the ability to infer the category and shape of objects around them on the road, such as other vehicles, road signs, pedestrians and obstacles. Different entities may have entirely different trajectories in the next moment. For example, a telephone pole is always stationary, while a pedestrian can move in arbitrary directions. Correctly determining the identity of different objects to deduce their motives is therefore crucial for motion planning in self-driving. Considering only the visible part of the object makes it more difficult to determine the category of the object, while reasoning about the full shape of the object helps to make more robust predictions \citep{qi2019amodal, breitenstein2022amodal}. Invisible areas of partially occluded objects are also prone to be missed or misjudged by a recognition algorithm. For example, during snowfall, obstacles on the road may be covered with snow and hence only partially visible. In this case, focusing only on the visible area of the object may trick an algorithm into seeing the snow-covered part of an obstacle as a passable road if they share features such as similar colour. Therefore, extending object predictions to the invisible portions of an object makes self-driving safer.

\subsection{Novel View Synthesis}
For a given image, novel view synthesis simulates the effect of having a virtual camera taking a new picture of the scene at a different angle of view. This application can reconstruct the environment depicted in the user's source image. The automatically-generated environment is intended to have a similar layout, content, and appearance to the source image. With several different novel views, users can feel as if they are walking freely in the scene of the original image. In addition to synthesising virtual views using typical methods such as disparity maps \citep{jampani2021slide} and depth maps \citep{hou2021novel}, image amodal completion techniques can be used to synthesise competitive novel views \citep{li20222d}. The advantage of the image amodal completion approach is that it offers a wider range of views from a single input image, rather than only being able to synthesise new views close to the original angle.

\section{OPEN ISSUES AND FUTURE DIRECTIONS}
There is no doubt that the existing approaches discussed above have contributed strongly to the development of the field of image amodal completion, but a number of challenges still lie ahead. In this section, we will discuss several promising future directions that are expected to further advance the field of image amodal completion.

\subsection{Challenges for Amodal Dataset}
The emergence of new datasets in the future will undoubtedly drive the field of image amodal completion. It is worth noting the challenges of different current data collection methods.

For real image datasets, the current ground-truth amodal masks are manually inferred by human annotators \citep{zhu2017semantic,qi2019amodal}. However, human predictions of the invisible parts of objects are subjective, and each individual potentially provides a different level of detail in their annotations. Therefore, one challenge is to make the amodal masks captured on the real images more closely resemble the actual ground truth. For instance, different imaging systems, such as X-rays, could be considered. In addition, a real dataset that provides visually complete ground truth about the appearance of the invisible parts of objects or backgrounds is currently unavailable. The next step could be to create real datasets with the ground-truth appearance in some simple scenes. For example, use a robotic arm to sequentially remove objects from the table.

Synthetic image datasets captured from computer-generated 3D scenes provide ground truth of amodal masks, amodal appearance and order information in detail by rendering. However, the current synthetic datasets are limited in terms of scenarios, with most focusing on indoor scenes. Future datasets are expected to cover more scenarios, including urban and natural environments.

Composite image datasets obtained by superimposing segments of images can also provide ground truth of amodal masks, amodal appearance and order information. Most of the current composite data are superimposed with visually dissimilar segments. Future datasets could use visually similar segments to trigger amodal completion tasks of similar objects occluding each other.

\subsection{Diversity of Outputs}

The diversity of hidden parts of objects is a natural question that accompanies image amodal completion. For an object that is only partially visible, the task of amodal completion is to predict a reasonable configuration of the hidden part, which has triggered different possibilities. For example, if only the upper half of a person is visible in the image, it is equally valid to infer whether the person is standing or sitting or in some other pose. Especially when most of an object is occluded, different reasonable assumptions could be made to fulfil the occluded region.

While most amodal appearance completion approaches tend to allow for multiple possible appearances using generative models, few studies on shape completion and order perception consider various possibilities. Due to the diversity of interpretations of the hidden parts of objects, one potential opportunity is to construct models that provide multiple reasonable patterns for each sub-problems of amodal completion. Alternatively, new methods may consider providing a single or diverse interpretation for invisible parts based on other aspects of the problem, such as the type of object involved.

It is worth noting that ground truth data containing only one solution can lead to difficulties in evaluating such probabilistic models. Creating multi-solution datasets or using metrics such as human observers' satisfaction and FID are expected to help compare the ability of models to predict multiple patterns.

\subsection{Better Visual Representations}

After generating the results of the amodal completion, a natural follow-up question is how to represent it. Currently, most of the amodal completion results inferred from individual RGB images are represented in 2D. However, how to present the additional inferred information in a more visual way is still to be investigated. One potential future approach is to build a 2.5D or 3D representation using additional information (such as the predicted layer order of the objects and the structure of the background) or using a hypothetical spatial layout (such as treating the background as multiple planes and the objects as simple volumes). Better visual representations not only provide a more intuitive interpretation of the outputs, but also encourage attention to additional information that potentially contributes to amodal completion problems, such as the layout of the space.

\subsection{Joint Resolution of Sub-problems}

Some studies \citep{ehsani2018segan, Papadopoulos_2019_CVPR, dhamo2019object, zhan2020self, zheng2021vinv} have attempted to address multiple sub-problems jointly, yet few studies explained the necessity of jointly solving these sub-problems. Therefore, a potential future work is to explain how important it is to solve the sub-problems of image amodal completion jointly. For example, in order to verify whether considering the pairwise order of occluded instance pairs contributes to the performance of the amodal shape (or appearance) prediction, an ablation study can be carried out to compare the results of experiments with and without pairwise orders. Explaining the importance of joint or isolated problem-solving will help researchers design more effective architectures for amodal completion problems.

\subsection{Consistent Performance Metrics}

As the image amodal completion community is still in the early stages of development, there are no consistent performance metrics for associated tasks at this time. As discussed in section \ref{metrics}, image amodal completion contains three tasks, and different studies for each task often adopt different evaluation metrics. Inconsistent evaluation metrics lead to difficulties in comparing multiple approaches for the same task. Therefore, consistent performance metrics for each image amodal completion task are required in the future.

\subsection{Real-time Models}

Despite developments in image amodal completion, computer vision systems are still a long way from being able to do this. For some applications, accuracy is the key issue, while for others it is vital to have models that can run in near real-time. A typical example is autonomous driving, which requires real-time reasoning about the invisible parts in order to make safe  driving plans. Robotic gripping systems also need immediate results in order to react quickly and get the job done efficiently. Most of the current models relating to image amodal completion pay little concern or omit to report model complexity and inference speed. Therefore, more efforts are needed for real-time amodal completion models in future research directions.

\subsection{Integration with Other Tasks or Contexts}

The integration of amodal completion tasks with other tasks or extra contexts holds great potential for improving the performance of the task. One future direction is to improve the flexibility and handling of uncertainty in amodal completion by utilising multi-modal completions \citep{wu2020multimodal, arora2022multimodal} that generate various representations. Additionally, the use of high-resolution completions \citep{han2017high} could provide more detailed results and better capture the nuances of the scene. Another promising direction is to integrate amodal completion with other tasks such as object detection, semantic segmentation, and depth estimation. Combining the strengths of multiple tasks can lead to more accurate and complete completions, as the model can leverage information from multiple sources to make more informed predictions. For example, combining with object detection provides better object instance information, while semantic segmentation provides richer background information. Furthermore, an important aspect to consider is the integration of context information, such as 3D information, temporal information, and scene context. These additional sources of information can provide the model with a better understanding of the objects and their environment, leading to more accurate results. Incorporating physical constraints such as object dynamics and physics could also improve the results, as they can provide the model with a more realistic understanding of the scene. The integration of these complementary tasks and information has the potential to provide a more comprehensive understanding of the scene, resulting in improved amodal completion performance.

\section{CONCLUSION}

The ability of amodal completion provides the computer vision system with information about the hidden parts of the images, including the whole shape of the objects (amodal shape completion), the complete visual appearance (amodal appearance completion) and the relationships between objects in the scene (order perception). The studies of image amodal completion allow computational models to behave more similarly to the human visual system and greatly benefit downstream tasks such as DR, self-driving and robotic grasping systems. However, in contrast to the extensive and mature work on the visible part of images, research on image amodal completion is still in its infancy.

To familiarise readers with recent developments in the field of image amodal completion, we provide a comprehensive survey of relevant methods to 2022. Three sub-problems are included in the main methods, each tending to take different approaches, with shape completion and order perception generally using discriminative models (assuming only one ground truth shape and correct ordering) and appearance completion typically using generative models (allowing many possible appearances). We discuss in detail the representative work for each sub-problem, along with their strengths and limitations. Then we detail the available datasets developed for image amodal completion, analysing key techniques for data collection and model evaluation.

Despite recent advances, there is still much room for improvement in current image amodal completion techniques, especially as practical applications requiring dealing with complex real-world scenarios remain challenging. Therefore, we provide some potential real-world applications, highlight open issues that require more attention, and suggest possible avenues for future research. We expect more interest in this emerging field, addressing current open issues and driving practical use in downstream applications.

\section*{Acknowledgments}
Jiayang Ao is supported by the Melbourne Research Scholarship.

\bibliographystyle{model2-names}
\bibliography{refs}

\end{document}